\documentclass[sigconf]{acmart}
\usepackage{subfigure}
\usepackage{enumitem}
\usepackage{multirow}
\usepackage[ruled,linesnumbered]{algorithm2e}
\usepackage{hyperref}
\usepackage{comment}
\usepackage{pifont}
\usepackage{amsmath}
\usepackage{graphics}
\usepackage{balance}

\AtBeginDocument{%
  }

\copyrightyear{2025}
\acmYear{2025}
\setcopyright{acmlicensed}\acmConference[KDD '25]{Proceedings of the 31st ACM SIGKDD Conference on Knowledge Discovery and Data Mining V.2}{August 3--7, 2025}{Toronto, ON, Canada}
\acmBooktitle{Proceedings of the 31st ACM SIGKDD Conference on Knowledge Discovery and Data Mining V.2 (KDD '25), August 3--7, 2025, Toronto, ON, Canada}
\acmDOI{10.1145/3711896.3736987}
\acmISBN{979-8-4007-1454-2/2025/08}

\begin{document}


\title{Gradients as an Action: Towards Communication-Efficient Federated Recommender Systems via Adaptive Action Sharing}


\author{Zhufeng Lu}
\email{10225101537@stu.ecnu.edu.cn}
\affiliation{%
  \institution{East China Normal University}
  \city{Shanghai}
  \country{China}
}

\author{Chentao Jia}
\email{51265902040@stu.ecnu.edu.cn}
\affiliation{%
  \institution{East China Normal University}
  \city{Shanghai}
  \country{China}
}

\author{Ming Hu}
\email{ecnu_hm@163.com}
\orcid{0000-0002-5058-4660}
\affiliation{%
  \institution{Singapore Management University}
  \city{Singapore}
  \country{Singapore}
}

\author{Xiaofei Xie}
\email{xfxie@smu.edu.sg}
\affiliation{%
  \institution{Singapore Management University}
  \city{Singapore}
  \country{Singapore}
}

\author{Mingsong Chen}
\email{mschen@sei.ecnu.edu.cn}
\affiliation{%
  \institution{East China Normal University}
  \city{Shanghai}
  \country{China}
}


\renewcommand{\shortauthors}{Zhufeng Lu, Chentao Jia, Ming Hu, Xiaofei Xie, and Mingsong Chen.}


\begin{abstract}

As a promising privacy-aware collaborative model training paradigm, Federated Learning (FL) is becoming popular in the design of distributed recommender systems.
However, Federated Recommender Systems (FedRecs) greatly suffer from two major problems: i) extremely high communication overhead due to massive item embeddings involved in recommendation systems, and ii) intolerably low training efficiency caused by the entanglement of both heterogeneous network environments and client devices.
Although existing methods attempt to employ various compression techniques to reduce communication overhead, due to the parameter errors introduced by model compression, they inevitably suffer from model performance degradation.
To simultaneously address the above problems, this paper presents a communication-efficient FedRec framework named FedRAS, which adopts an action-sharing strategy to cluster the gradients of item embedding into a specific number of model updating actions for communication rather than directly compressing the item embeddings.
In this way, the cloud server can use the limited actions from clients to update all the items.
Since gradient values are significantly smaller than item embeddings, constraining the directions of gradients (i.e., the action space) introduces smaller errors compared to compressing the entire item embedding matrix into a reduced space.
To accommodate heterogeneous devices and network environments, FedRAS incorporates an adaptive clustering mechanism that dynamically adjusts the number of actions.
Comprehensive experiments on well-known datasets demonstrate that FedRAS can reduce the size of communication payloads by up to 96.88\%, while not sacrificing recommendation performance within various heterogeneous scenarios. We have open-sourced FedRAS at https://github.com/mastlab-T3S/FedRAS.

\end{abstract}



\maketitle



\section{Introduction}
Recommender systems~\cite{resnick1997recommender, lu2012recommender, barkan2024counterfactual, jung2023dual}, which take advantage of user-item behavior data to provide effective item recommendations, have been widely adopted in advertising~\cite{GARCIASANCHEZ2020102153} and e-commerce~\cite{4280214} applications.
However, growing concerns about data privacy breaches and stringent regulations imposed by major economies worldwide~\cite{bonta2022california, voigt2017eu} have made it increasingly challenging to collect high-quality user data for model training of recommender systems.
Federated Learning (FL)~\cite{fedavg,huang2024federated,hu2024fedmut,qi2023cross,hu2024aggregation,xia2025multisfl,liao2024swiss, li2024joint, yang2024improved}, a distributed machine learning paradigm that preserves privacy, enables training of collaborative models without data sharing. 
Based on FL, various Federated Recommender Systems (FedRecs)~\cite{muhammad2020fedfast, wang2022fast, yang2020federated, wang2021poi} have been proposed to address data privacy concerns in recommendation tasks.

Although FedRecs offers a promising solution to alleviate the risk of privacy leakage, it faces two critical challenges: \ding{182} extremely high communication overhead due to the massive number of items in recommendation models, and \ding{183} significantly low training efficiency caused by heterogeneous network environments and device hardware limitations.
Specifically, the parameters of a recommendation model typically consist of user and item embeddings, where the number of item embeddings corresponds to the total number of items in the target recommender system. Since recommender systems often involve a massive number of items, FedRecs inevitably incur substantial communication overhead for model transmission between the server and devices.
In addition, since recommendation systems are typically deployed on devices with limited hardware resources, such as mobile phones and personal computers, excessive communication overhead leads to significant degradation in training performance~\cite{hu2023gitfl, cui2022optimizing, wang2023dafkd}.
Furthermore, the heterogeneous network bandwidth across devices and unstable network environments further exacerbate this performance degradation~\cite{flexfl,gao2024nebulafl}.

To improve communication efficiency in FedRecs, existing methods attempt to compress item embeddings using strategies such as meta-learning~\cite{FedMF}, matrix factorization~\cite{JointRec, CoLR}, or row reduction~\cite{FCF-BST}. 
Although these methods effectively reduce communication overhead, lossy compression of model parameters inevitably leads to model performance degradation. 
Specifically, these methods directly compress item embeddings for transmission between the cloud server and devices. 
Due to the lossy nature of compression, the decompressed embeddings significantly deviate from the original ones, resulting in a notable decline in model performance.
Therefore, \textit{achieving communication-efficient FedRecs while maintaining model performance remains a critical and unresolved challenge in the design of FedRecs}.

Typically, gradient values are much smaller than the embeddings themselves.
Intuitively, compared to directly compressing the embeddings, compressing gradients may result in relatively smaller compression loss.
From a geometric perspective, similar to controlling the direction of a vehicle's movement, the gradient can be viewed as an action that moves the embedding vector a certain distance in a specific direction.
Typically, even though the value of gradients is changed slightly, the updated gradients still have a high probability of remaining aligned with the direction of gradient descent.
Moreover, as the training continues, subsequent training can more easily correct the deviation in the optimization direction caused by gradient compression.
Based on the above intuition, to limit the communication overhead, we can limit the number of actions, where items with similar gradients share the same approximate gradients.
In this way, FedRec only needs to transfer a specific number of actions and the action index for each item instead of the entire item embedding matrix across the cloud server and devices.
Note that the cloud server and devices can update their maintained item embeddings with received actions and corresponding indies.

Inspired by the above intuition, to achieve high-performance and communication-efficient FedRec, this paper presents a novel FedRec framework named \textit{FedRAS}, which adopts an adaptive action-sharing strategy for communication across the cloud server and devices.
Specifically, FedRAS transfers actions, i.e., gradients rather than item embeddings across the cloud server and devices.
To reduce communication overhead, FedRAS clusters the gradients of items to a specific number of groups and adopts the center of each group to approximate the gradients in the corresponding group, where these approximated gradients are used for communication and item embedding updating.
To accommodate heterogeneous network environments and device resources, FedRAS utilizes an adaptive clustering mechanism to dynamically adjust the number of groups according to the target group number and similarity of gradients within a group, which simultaneously resolves the challenge of cross-round gradient distribution variability.
In summary, this paper makes three main contributions as follows:
\begin{itemize}

\item[$\bullet$] We present a novel communication-efficient FedRec framework named FedRAS, which clusters the gradients of the items to a specific number of actions rather than all the item embeddings for communication.

\item[$\bullet$] We propose an adaptive clustering mechanism to adaptively adjust the number of actions based on the similarity of gradients and the heterogeneity of devices and environments.

\item[$\bullet$] We conduct extensive experiments on various well-known datasets to demonstrate the effectiveness and efficiency of our FedRAS approach.

\end{itemize}

The remainder of this paper is organized as follows. 
Section 2 introduces the preliminary knowledge and related work. 
Section 3 discusses our motivation and ideas. 
Section 4 details our FedRAS approach together with its key components.  
Section 5 presents the experimental results.  
Finally, Section 6 concludes the paper.


\section{Preliminary and Related Work}

\subsection{Preliminary}
\textbf{FL-based Recommendation.} In this work, we focus on a scenario where recommendations are derived exclusively from the user-item interaction matrix without incorporating additional user or item attributes. There are $M$ users and $N$ items where each user has a private interaction set $O_u = \{(i, r_{iu})\} \subseteq [N] \times \mathbb{R}$. Each user is treated as an individual client. The goal of FedRecs is to minimize the aggregated loss function
$\underset{\{\mathbf{u}_{1,\ldots,M},\mathbf{Q},\mathbf{\theta}\}}{\operatorname*{argmin}}\sum_{u=1}^{M}\mathcal{L}(\mathcal{F}(\mathbf{p}_{u_i},\mathbf{Q},\mathbf{\theta})),
$
where $\mathbf{Q}$ denotes the item embeddings, $\mathbf{p}_{u_i}$ denotes the user embedding of client $i$, $\theta$ is the scoring model parameter and $\mathcal{F}$ represents the recommendation algorithm. 
Each client optimizes its local model using the following binary cross-entropy loss function: 
$$
    \mathcal{L}(\mathcal{F}(\mathbf{u}_i,\mathbf{Q},\mathbf{\theta})) = \sum_{(i, r_{iu})\in{O_u}} r_{iu}\mathrm{log}(\hat{r}_{iu})+(1-r_{iu})\log(1-\hat{r}_{iu})),
$$
where $r_{ij}$ is the target, and $\hat{r}_{ij}$ represents the predicted results. In implicit recommendation settings, $r_{ui} = 1$ denotes a positive interaction between user $u$ and item $i$, and vice versa.

\textbf{Clustering Method.}
We use the K-means algorithm~\cite{kmeans} as the clustering method. 
Given a data matrix $Q \in \mathbb{R}^{N \times d}$, the target is to cluster the rows of $Q$ into $K$ groups, i.e., to assign the row index of $Q(i,:)$ to the group $G_k$, $k \in \{1, \ldots, K\}$. The target of K-means can be written in optimization form as follows:
$$
\operatorname*{argmin}_{G_1, \dots, G_K} \quad \sum_{k=1}^{K} \sum_{i \in G_k} \| Q(i, :) - \mu_k \|_2^2 ,
$$
where $\mu_k$ is the centroid of group $G_k$, i.e., $\mu_k = \frac{1}{|G_k|} \sum_{i \in G_k} Q(i, :)$.

\subsection{Related Work}

\textbf{Federated Recommender Systems.}
In recent years, based on the architecture of FL~\cite{fedavg,qi2023cross,hu2024fedcross,xia2024cabafl}, FedRecs have received a great deal of attention due to their excellent privacy protection capabilities. FCF~\cite{FCF} and FedRec~\cite{FedRec} are the pioneering federated recommendation frameworks to predict ratings based on matrix factorization. 
FedMF~\cite{FedMF} introduced homomorphic encryption (HE) technology based on non-negative matrix encryption gradient to achieve privacy protection.
Lin \textit{et al.} proposed MetaMF~\cite{MetaMF}, which uses a meta-network to generate rating prediction models and private item embedding. FedNCF~\cite{FedNCF} adapted Neural Collaborative Filtering (NCF)~\cite{NCF} to the federated setting, incorporating neural networks to learn user-item interaction functions. ~\cite{FedPerGNN} presented FedPerGNN, where each user maintained a Graph Neural Network (GNN) model to incorporate high-order user-item information.

\textbf{Communication Efficient Federated Recommendation.}
In FedRecs, the large item embedding layer incurs significant communication overhead, which greatly impacts the training efficiency of federated learning. 
To address this issue, three promising directions have been explored: \textit{meta-learning techniques}, \textit{matrix factorization}, and \textit{row reduction}.
Meta-learning technology~\cite{MetaMF} effectively avoids the communication overhead by reducing the entire item embedding layer.
Low-rank matrix factorization is seen as a promising approach to reducing dimensionality. For example, JointRec~\cite{JointRec} reduced uplink costs by using low-rank matrix factorization and 8-bit probabilistic quantization to compress the item matrix. CoLR~\cite{CoLR} used low-rank matrix factorization to divide payloads into a fixed part and a transmitter part. 
Another solution focused on reducing the number of items transmitted downlink, thereby reducing the number of rows in the item matrix. It used a multi-arm bandit algorithm to address item-dependent payloads, which was proposed by Khan et al.~\cite{FCF-BST}.
However, existing methods fail to maintain the performance of the model while realizing communication optimization, especially under extreme low-quality communication bandwidth scenarios. In addition, the heterogeneous network
environments and device resources of real-world FL training process make it difficult for these methods to be applied.

To the best of our knowledge, FedRAS is the first attempt to utilize a cluster-based strategy to solve the payload optimization problem for FedRecs. Using an action sharing strategy, FedRAS can greatly reduce communication overhead without affecting performance.
To deal with heterogeneous communication bandwidth scenarios, we explore a novel adaptive clustering mechanism, which can adaptively adjust the clustering granularity according to the clients' communication bandwidth and effectively improve the performance of FL training.

\section{Motivation}
Typically, FedRecs compresses the model parameters to reduce the communication overhead.
Since item embeddings account for most of the model parameters, existing methods primarily focus on compressing item embeddings.
However, directly compressing the item embeddings leads to the destruction of item features, inevitably resulting in degradation of the model performance.

\textbf{Intuition.}
From a geometric perspective, each item embedding can be represented as an $n$-dimensional vector, corresponding to a point in $n$-dimensional space.
The gradients of an item embedding can then be viewed as the displacement of that point in the space.
Typically, the magnitude of gradients is much smaller than that of the item embeddings themselves.
Intuitively, compressing gradients, rather than item embeddings, can mitigate the distortion of item features and significantly reduce errors.
Since both the cloud server and devices maintain complete copies of all item embeddings, FedRec can transfer gradients instead of full item embeddings during FL training. 
Upon receiving the gradients, the cloud server or devices can update their item embeddings by adding the received gradients.
Therefore, compressing the gradients may be a promising strategy for communication-efficient FedRec.

\textbf{Our Ideas.} 
Based on the above intuition, we attempt to compress the gradients rather than the item embeddings.
Typically, the direction of gradients indicates the direction of loss descent for the corresponding item embedding.
Intuitively, since gradient descent is inherently stochastic, even a slight change in the gradient direction still ensures movement toward decreasing the loss.
Therefore, we can approximate the gradients using a set of representative gradients that satisfy a predefined similarity threshold.
We can interpret the gradients of item embeddings as actions for updating, and the item embeddings with similar gradients can use the same action to update.
Based on this insight, we attempt to use a fixed number of actions to approximate the gradients of all the item embeddings.
Specifically, we cluster the gradients of all the item embeddings to a specific number of groups and adopt the center of the group as the action.
In this way, FedRec only transfers a specific number of actions across the cloud server and devices for model updating rather than the gradients of all the item embeddings, significantly reducing communication overhead.





\begin{figure*}[!t]
\centering
\includegraphics[width=0.85\linewidth]{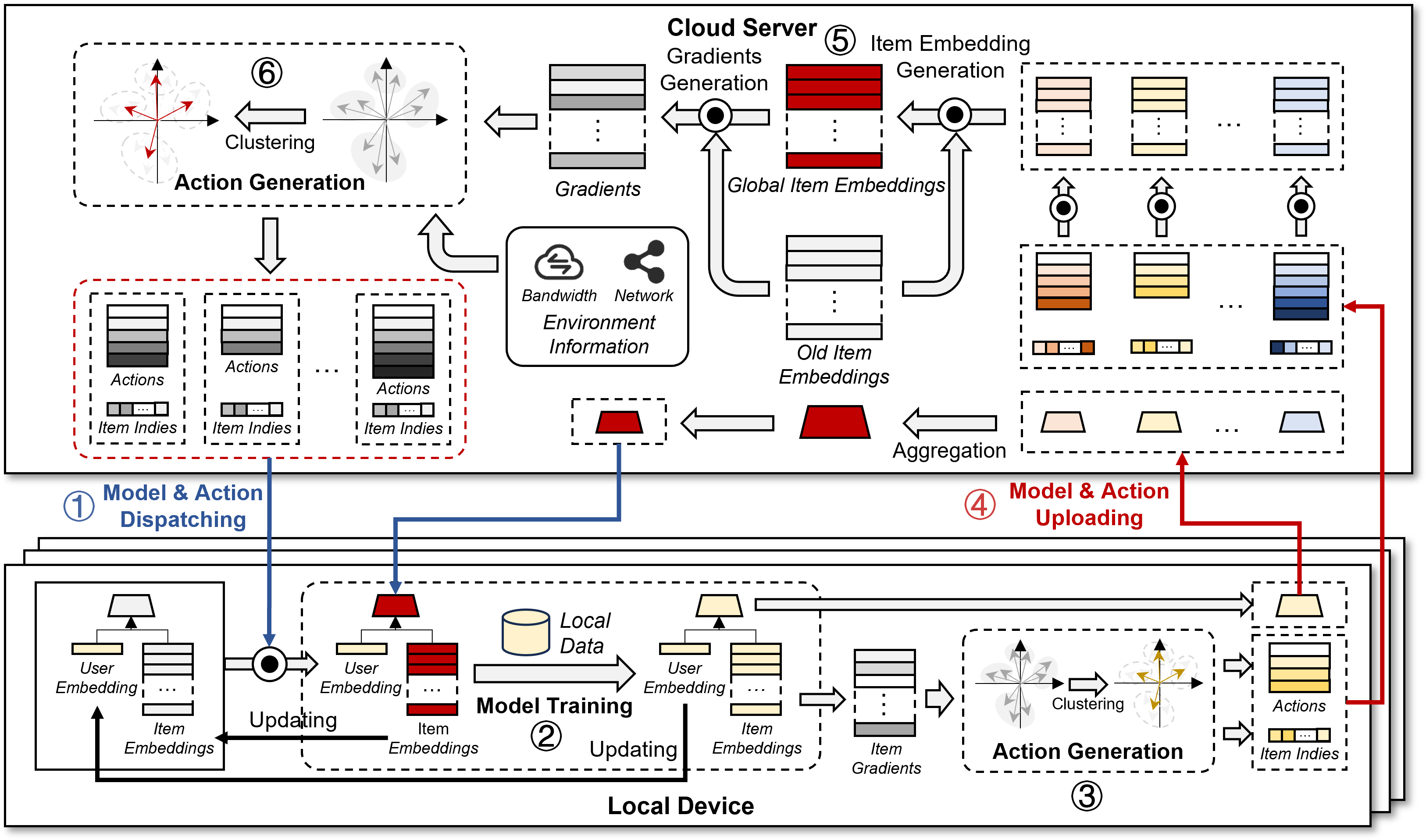}
\caption{Framework and workflow of FedRAS.}
\label{fig: framework}
\end{figure*}

\section{Our FedRAS Approach}

Figure \ref{fig: framework} presents the framework and workflow of FedRAS. As shown in the figure, FedRAS uses an action-sharing strategy to reduce the communication load between the server and clients. 
To accommodate heterogeneous network environments and device resources, it employs an adaptive clustering mechanism on the server to generate action sets of various sizes. 
Additionally, to improve clustering effectiveness and accelerate convergence, FedRAS employs a new global model generation mechanism to obtain more accurate gradients.

Specifically, each training round of FedRAS consists of six steps:
\begin{itemize}
    \item \textbf{Step 1: Model \& Action Dispatching.} 
    The server sends the aggregated scoring model from the previous round, the actions (compressed item gradients), and the action index for each item to the participating clients based on their bandwidth quality and communication resources.
    \item \textbf{Step 2: Model Training.} 
    Each client updates its local item embeddings with the received actions and then trains the model using its own data.
    \item \textbf{Step 3: Action Generation.} 
    The client determines whether gradient compression is needed. If the size of the item gradients to be uploaded exceeds the communication overhead threshold, the client compresses them using clustering.
    \item \textbf{Step 4: Model \& Action Uploading.} 
    The client uploads the trained scoring model and actions to the server.
    \item \textbf{Step 5: Global Model Generation.} 
    After receiving models and actions from all activated clients, the server reconstructs the original gradients of items and aggregates them to obtain the updated item embeddings and scoring model.
    \item \textbf{Step 6: Adaptive Clustering.} 
    The server uses an adaptive clustering mechanism to compress the gradients and generates action sets of various sizes for heterogeneous clients based on environment information.
\end{itemize}


\subsection{Implementation of FedRAS}

\begin{algorithm}[h]
\caption{Implementation of FedRAS}
\label{alg: FedRAS}
\SetKwFunction{ClientOpt}{ClientOpt}
\SetKwFunction{Kmeans}{KmeansCompress}
\SetKwFunction{Ada}{AdaCompress}
\SetKwFunction{Aggregation}{Aggregation}
\SetKwFunction{Localtrain}{ClientLocalTrain}
\SetKwFunction{DownloadParam}{DownloadParam}
\SetKwFunction{ClientSelect}{ClientSelect}
\SetKwFunction{NonzeroRowCount}{NonzeroRowCount}
\SetKwFunction{Reconstruct}{Reconstruct}
\KwIn{
    i) Client set $C$; 
    ii) Global epoch $T$; 
    iii) Local epoch $\tau$; 
    iv) Expected group number $C_e$; 
    v) Clustering fluctuation factor $\alpha$;
    vi) Bandwidth limitation of all clients $E=\{e_1,\dots,e_{|C|}\}$.
}
\SetKwProg{Fn}{Function}{:}{}

Initial item embedding $Q^{(0)}$ and scoring model $\theta^{(0)}$

\For{$t \in \{0, 1, 2, \dots, T\}$}{
    $S^{(t)} \gets$ \ClientSelect($C$)
    
    /* parallel For */ \\
    \For{$u \in S^{(t)}$}{
        $\Delta Q_{c_u}^{(t)}, \theta_u^{(t)}  \gets  $ \Localtrain($u$)\; 
    }
    
    $\Delta Q^{(t)} \gets$ \Aggregation($\{\Delta 
    Q_{c_u}^{(t)}\mid \forall u \in S^{(t)}\}$)\;

    $Q^{(t)} = Q^{(t-1)} + \Delta Q^{(t)}$\;
    
    $\theta^{(t)} \gets \frac{1}{|S^{(t)}|} \sum_{u}^{S^{(t)}} \theta_u^{(t)}$\;
    
    $\{\Delta Q_{c_1}^{(t+1)}, \dots, \Delta Q_{c_{|C|}}^{(t+1)}\} \gets$ \Ada($\Delta Q^{(t)}, C_e, \alpha, E$)\;
}

\Fn{\Localtrain{$u$}}{
    \If{$t > 0$}{
     $\Delta Q_{c_u}^{(t)}, \theta^{(t)} \gets $\DownloadParam($t, u$)\;
        $Q_u^{(t)} \gets Q_u^{(t-1)} +$ \Reconstruct($\Delta Q_{c_u}^{(t)}$)\;
        $\theta_u^{(t)} \gets \theta^{(t)}$\;
    }
    
     $Q_u^{(t, 0)} \gets Q_u^{(t)}$, $\theta_u^{(t,0)} \gets \theta^{(t)}$\;
    $\Theta_u^{(t, 0)} \gets \{ p_u^{(t, 0)}, Q_u^{(t, 0)}, \theta_u^{(t, 0)} \}$\;
    
    \For{$k \in \{0, 1, \dots, \tau - 1\}$}{
        $\Theta_u^{(t, k+1)} \gets$ \ClientOpt($\Theta_u^{(t, k)}$)\;
    }
    
    $p_u^{(t+1)} \gets p_u^{(t, \tau)}$, $\theta_u^{(t+1)} \gets \theta_u^{(t, \tau)}$\;
    $\Delta Q_u^{(t)} \gets Q_u^{(t, \tau)} - Q_u^{(t)}$\;
    
    \If{\NonzeroRowCount($\Delta Q_u^{(t)}$) > $e_u$}{
        $\Delta Q_{c_u}^{(t)} \gets$ \Kmeans($\Delta Q_u^{(t)}$, $e_u$)\;
    }
    
    \Return  $\Delta Q_{c_u}^{(t)}$, $\theta_u^{(t+1)}$\;
}
\end{algorithm}

Algorithm \ref{alg: FedRAS} details the implementation of FedRAS. 
Before FL training, line 1 initializes the global model parameters. Lines 2-11 detail the training process for each round. In line 3, a random subset of clients is selected to participate in the current training round. Lines 5-7 represent the local training process performed by participating clients in parallel.
Lines 13-29 detail the function ClientLocalTrain, which describes the process of local training. Specifically, the client $u$ first downloads the compressed gradient matrix $\Delta Q_{c_u}$ from the server and uses it to compute the new item embedding matrix. Lines 21-23 explain that the client performs multiple rounds of local training. The term ClientOpt denotes the optimization method, i.e., gradient descent, used by the client to update the parameters. In line 25, the client calculates the gradient matrix based on the difference between the model parameters before and after training. Lines 26-28 indicate that if the number of non-zero rows in the client's gradient matrix exceeds its bandwidth limitation $e_u$, the client applies K-means clustering for gradient compression. Note that this bandwidth limitation is measured by the number of rows in the item embedding.
Back to the server-side algorithm, lines 8 and 9 indicate that the server aggregates gradients and generates the new item embedding. In line 11, the server applies our proposed adaptive clustering mechanism to perform gradient compression, generating multiple compressed gradient matrices for the next round of training.

\subsection{Action Sharing}

In order to reduce the cost of communication, FedRAS uses clustering to compress the aggregated gradients of items. 
We use the K-means algorithm to obtain the clustering results and then use the centroid vector of each group to encode the gradients in the corresponding group. 
Assume that the action (gradient) of the $i$-th item is denoted as $\Delta q_i$, and it belongs to group $G_k$. It can be encoded as
$
    \Delta q_i^{Enc} = \text{Enc}(\Delta q_i) = \mu_k,
$
where $\mu_k$ represents the centroid of group $G_k$, and \text{Enc}(·) denotes the action sharing algorithm using K-means clustering.

Once the server has encoded the gradients for all items, it can compress the item gradient matrix $\Delta Q$ into a compressed matrix $\Delta Q_c$ consisting of group centroids, while also recording the group index for each item. 
The server then sends the group centroids and group indices to the clients involved in the current round of training. 
Based on the group index information, clients can easily reconstruct the full item gradient matrix.

In addition to reducing downlink communication costs, this cluster-based compression method is also used on clients. 
Since each client only trains a subset of the items in each round, and the updates for untrained items are typically zero, only the updates for the items trained in the current round need to be uploaded. 
Based on this observation, FedRAS employs a selectable compression strategy, where gradients are compressed if the size of the gradient matrix to be uploaded by the client exceeds a certain threshold $e_u$ based on its communication resources; Otherwise, the gradients are directly uploaded without compression.

\subsection{Adaptive Clustering Mechanism}
To accommodate heterogeneous network environments and varying gradient distributions across rounds, we proposed an adaptive clustering mechanism. This mechanism employs an iterative cluster-and-split algorithm to dynamically determine the number of groups in each round and generate multiple compressed gradient matrices. 
An automatic cosine similarity calculation algorithm is used to set the stopping condition. 

{\bf Cluster-and-split Algorithm.} 
The process begins by clustering all gradients, followed by finding groups with poor clustering quality. 
These groups are then iteratively split in a binary fashion until the stopping condition is satisfied. 
A fluctuation factor $\alpha\in(0, 1)$ is used to limit the range of groups for all rounds. 
Suppose that the expected number of groups $C_e$ has been set to meet the preset communication compression rate. 
We can calculate the min (max) group number $C_i$ ($C_m$) using the equations $C_i = C_e * (1 - \alpha)$ and $C_m = C_e * (1 + \alpha)$.
At the beginning of this approach, all gradients to be transmitted are clustered into $C_i$ groups. 

Since the gradient direction has a great impact on the convergence speed, we use the average cosine similarity between all gradients in a group and the group center as the metric to measure the clustering quality. 
The average cosine similarity of a group can be obtained by the equation:
$$
    \text{AvgCosSim}(c)
= \frac{1}{|c|}\sum_{i \in c} \cos(\Delta q_i^c, \mu_c)
= \frac{1}{|c|}\sum_{i \in c} \frac{\Delta q_i^c \cdot \mu_c}{\|\Delta q_i^c\|\|\mu_c\|} , 
$$
where cos(·, ·) is the cosine similarity function, $\Delta q_i^c$ denotes the gradient of the $i$-th item in group $c$, $\mu_c$ denotes the center of $c$, and $|c|$ denotes the number of elements in $c$. 
The group with the smallest average cosine similarity is chosen for splitting.

We use a binary splitting strategy as the splitting algorithm. 
Firstly, the cosine similarity matrix of the gradients in a group is calculated to find two gradients with the lowest cosine similarity. 
Then, these two gradients are taken as the centers of two new groups, and all the gradients are assigned to the two groups according to the magnitude of the cosine similarity with the two group centers. 
This algorithm can be described as follows: 
$$
    (\mu_{1}, \mu_{2}) = \mathop{\arg\min}\limits_{(\Delta q_i, \Delta q_j),\, i \neq j} \cos(\Delta q_i, \Delta q_j), 
$$
$$
C_{1} = \{\Delta q \in C \;|\; \cos(\Delta q, \mu_{1}) \;\ge\; \cos(\Delta q, \mu_{2})\},
$$
$$
C_{2} = \{\Delta q \in C \;|\; \cos(\Delta q, \mu_{2}) \;>\; \cos(\Delta q, \mu_{1})\},
$$
where $\mu_{1}$ and $\mu_{2}$ denote the two gradients with the lowest cosine similarity in $C$,  $C$ denotes the original group of gradients, and $C_{1}$ and $C_{2}$ denote the two newly formed sub-groups. 
As the splitting process progresses, multiple compressed gradient matrices can be obtained. 
The splitting process is terminated when the minimum average cosine similarity among all groups falls below a certain threshold or when the maximum number of groups $C_m$ is reached. 

{\bf Automated Calculation for Cosine Similarity Threshold.} 
Directly setting the cosine similarity threshold as a hyperparameter and tuning it through the validation set is challenging, and the communication overhead is hard to control. 
Therefore, we designed an algorithm to automatically calculate the cosine similarity threshold using information from all previous rounds.
Specifically, when the number of groups reaches the desired value $C_e$ during the splitting process of each round, the minimum average cosine similarity of all groups is recorded. 
The threshold for the current round is then defined as the average of all recorded values from previous rounds. 
For the initial round (i.e., round 0), the groups are directly split to $C_e$. 
For the other round, if the current number of groups is less than $C_e$ and the splitting process stops, the current splitting result is recorded. 
The splitting process then continues until the number of groups reaches $C_e$. 
The server will use the splitting result from the first stop to compress the gradient matrix.

\subsection{Global model generation}
After receiving the models and actions from the clients, the server needs to reconstruct and aggregate them to generate a new model. 
In FedRecs, the model on the server typically consists of two parts, i.e., the item embedding and the scoring model.

{\bf Item Embedding Generation.}
To obtain more accurate gradients for improved clustering effectiveness and accelerated convergence, we adopted a new gradient aggregation strategy. 
Specifically, the server will calculate the number of clients involved in the training of each item. Let $S_t$ denotes the number of clients activated in round $t$, and $S_t^{i} \in S_t$ represent the set of clients that participate in the training of item $i$. Then, the new aggregation strategy for the gradient of item $i$ can be expressed as follows:
$$
    \Delta q_i = \frac{1}{|S_t|}  \sum_{u=1}^{S_t}\Delta q_{i}^{u} \times \frac{|S_t|}{|S_i|} .
$$
Next, the aggregated gradient matrix is used to update the item embeddings on the server:
$$
    Q^{(t)} = Q^{(t-1)} + \Delta Q^{(t)} . 
$$

{\bf Scoring Model Generation.} 
The server uses FedAvg to aggregate the scoring models uploaded by the clients participating in the current training round. This process can be described as:
$$
    \theta^{(t)} = \frac{1}{|S_t|} \sum_{u}^{S_t} \theta_u^{(t)}.
$$

\begin{table*}[!h]
\small
\centering
\caption{Metric value (HR@10/NDCG@10) of models under different compression rates (CR). The best results are shown in \textbf{bold}.}

\label{Table: main exp}
\setlength{\tabcolsep}{1.7pt}
\begin{tabular}{cc|ccc|ccc|ccc}
\hline
\multirow{2}{*}{Model} & \multirow{2}{*}{Method} & \multicolumn{3}{c|}{MovieLens-100K} & \multicolumn{3}{c|}{MovieLens-1M} & \multicolumn{3}{c}{Lastfm-2K} \\ \cline{3-11}
 &  & CR=90.63\% & CR=93.75\% & CR=96.88\% & CR=90.63\% & CR=93.75\% & CR=96.88\% & CR=90.63\% & CR=93.75\% & CR=96.88\% \\ \hline
\multirow{5}{*}{MF} 
& Base   & 0.5027/0.2625 & 0.3934/0.2117 & 0.3892/0.1990 & 0.4562/0.2526 & 0.4579/0.2529 & 0.4551/0.2503 & 0.6162/0.4159 & 0.5944/0.3973 & 0.4950/0.3284 \\
& TopK   & 0.3722/0.1988 & 0.3415/0.1835 & 0.3309/0.1772 & 0.3733/0.2085 & 0.3399/0.1929 & 0.3969/0.2142 & 0.6837/0.5282 & 0.6875/0.5229 & 0.6631/0.5020 \\
& SVD    & 0.6013/0.3228 & 0.5854/0.3158 & 0.5228/0.2745 & 0.5902/0.3262 & 0.5667/0.3121 & 0.4904/0.2702 & 0.7725/0.5724 & 0.7625/0.5608 & 0.7156/0.5267 \\
& CoLR   & 0.5992/0.3225 & 0.5960/0.3228 & 0.5589/0.3008 & 0.6124/0.3443 & 0.5873/0.3286 & 0.5710/0.3164 & 0.7606/0.5619 & 0.7538/0.5545 & 0.7344/0.5399 \\
& FedRAS & \textbf{0.6341/0.3540} & \textbf{0.6331/0.3499} & \textbf{0.6299/0.3459} & \textbf{0.6632/0.3852} & \textbf{0.6581/0.3829} & \textbf{0.6496/0.3711} & \textbf{0.8137/0.6843} & \textbf{0.8194/0.6948} & \textbf{0.8194/0.6918} \\ \hline
\multirow{5}{*}{NCF} 
& Base   & 0.5164/0.2202 & 0.2874/0.0928 & 0.00/0.00      & 0.4743/0.2480 & 0.4550/0.2511 & 0.00/0.00      & 0.7475/0.4674 & 0.7350/0.4598 & 0.00/0.00      \\
& TopK   & 0.3913/0.2114 & 0.4040/0.2069 & 0.3839/0.2033 & 0.4401/0.2429 & 0.4490/0.2430 & 0.4455/0.2433 & 0.6225/0.4179 & 0.6094/0.3993 & 0.6350/0.4237 \\
& SVD    & 0.5790/0.3077 & 0.5705/0.2980 & 0.5345/0.2836 & 0.5474/0.3013 & 0.5435/0.3029 & 0.4402/0.2382 & 0.7575/0.5393 & 0.7462/0.5327 & 0.6787/0.4721 \\
& CoLR   & 0.5779/0.3095 & 0.5599/0.2940 & 0.5514/0.2876 & 0.5599/0.3021 & 0.5305/0.2847 & 0.5068/0.2769 & 0.6900/0.4821 & 0.6900/0.4856 & 0.6825/0.4660 \\
& FedRAS & \textbf{0.5832/0.3116} & \textbf{0.5832/0.3164} & \textbf{0.5716/0.3072} & \textbf{0.5977/0.3331} & \textbf{0.5844/0.3264} & \textbf{0.5757/0.3198} & \textbf{0.8225/0.6269} & \textbf{0.8200/0.6286} & \textbf{0.8150/0.6162} \\ \hline
\end{tabular}
\end{table*}

\section{Experimental Results}






In this section, we conducted experiments to validate the effectiveness of our approach and further analyze its properties. All experimental results were obtained from an Ubuntu workstation equipped with an Intel Xeon Platinum 8336C CPU, 128GB memory, and eight NVIDIA GeForce GTX-4090 GPUs. 

\subsection{Experimental Settings}

{\bf Datasets}.
We conducted our experiments on three datasets, i.e.,  MovieLens-100K, MovieLens-1M~\cite{ML}, and Lastfm-2K~\cite{Lastfm}, which are widely used to evaluate recommendation models. 
We excluded users with less than $5$ interactions in Lastfm-2K. The detailed dataset information is shown in Table~\ref{tab: datasets}. For dataset split, we followed the prevalent leave-one-out evaluation~\cite{NCF}. We evaluated the model performance with two important metrics, i.e., Hit Ratio (HR) and Normalized Discounted Cumulative Gain (NDCG).

\begin{table}[H]
\caption{Dataset statistics.}
\label{tab: datasets}
\small
\begin{tabular}{ccccc}
\toprule
Dataset&Interactions&Users&Items&Sparsity\\
\midrule
MovieLens-100K & 100,000 & 943 & 1,682 & 93.70\%\\
MovieLens-1M & 1,000,209 & 6,040 & 3,706 & 95.53\%\\
Lastfm-2K & 70,390 & 1,482 & 12,399 & 99.62\%\\
\bottomrule
\end{tabular}
\end{table}

{\bf Base Models}.
We chose two classical models, i.e., MF and NCF~\cite{NCF}, as the backbone in our experiments. MF calculates the similarity by taking the dot product between the user embedding and the item embedding, while NCF replaces the dot product operation with a neural network. In this work, the neural network used three feedforward layers with 64, 32, and 16 dimensions, respectively.

{\bf Baselines}.
To evaluate the effectiveness of FedRAS, we compared our approach with multiple communication optimization methods of FedRecs as follows:
\begin{itemize}

\item[$\bullet$] \textbf{Base}. The classical FL algorithm FedAvg~\cite{fedavg} in FedRecs. For the MF model, ``Base'' stands for FedMF~\cite{FedMF}; For the NCF model, ``Base'' stands for FedNCF~\cite{FedNCF}. We simply reduced the dimensions of all embedding vectors to accommodate different compression rates.

\item[$\bullet$] \textbf{Topk}. This method employs sparsification, representing updates as sparse matrices to reduce communication costs. Specifically, only the top-k gradient values (per row) in the embedding matrix are transmitted between the client and server, while the rest are pruned.

\item[$\bullet$] \textbf{SVD}. This method uses a low-rank structure to compress the communication load, which is based on singular value decomposition (SVD). 

\item[$\bullet$] \textbf{CoLR}~\cite{CoLR}. CoLR is a SOTA communication-efficient federated recommendation framework that decomposes the gradient matrix into the product of two submatrices and reduces communication overhead by fixing one submatrix while aggregating the other.

\end{itemize}

{\bf Hyperparameter Settings}.
To ensure fairness, we adopt SGD as the optimizer, and the original user/item embedding size for all methods defaults to 32. 
For each dataset, we set the number of clients participating in each round to be equal to 10\% of the number of all users. 
For local training, we sampled four negative instances per positive instance guided by ~\cite{NCF}, set the batch size to 256 and the local epoch to 2.
For FedRAS, we set the target number of groups $C_e$ for gradient clustering according to the preset communication overload. The clustering fluctuation factor $\alpha$ is set to 0.2 by default. 

{\bf Simulation of Communication-Constrained Scenarios}.
To simulate scenarios with limited communication resources, we use the compression rate as an indicator to measure communication overhead. 
The compression rate (CR) is calculated as $CR=1-cost_{current}/cost_{original}$, where $cost_{original}$ refers to the size of the entire embedding matrix. 
The expected group count $C_e$ is calculated by $N*(1-CR)$, where $N$ denotes the number of items. For a fair comparison with the baselines, we treated all clients equally without considering heterogeneity, which means that $e_u$ is equal to $C_e$ for each user $u$.
For each method, we perform compression on both uplink and downlink communication loads. 
In our FedRAS approach, in addition to transmitting the clustered gradient centroids, we also communicate the group assignment indices for the gradient of each item. Since these indices can be encoded with far fewer bits (compared to the 32-bit floating-point representation of gradients), their contribution to the total communication overhead is negligible. Thus, we omit this term in our final overhead calculation. 
Note that the actual compression rate of FedRAS will be slightly different from the preset one due to the adaptive clustering mechanism.

\subsection{Performance Comparison}
\label{sec: expr}
We evaluated the performance of FedRAS and four baselines on two base models with different compression rates. 

{\bf Comparison of Model Performance}.
Table~\ref{Table: main exp} compares the performance between FedRAS and all baselines on three datasets. In this table, the first column indicates the type of base model. The second column denotes different methods to compress the communication overload. The remaining columns show the performance of each method on the three datasets under different compression rates. 
From this table, we can find that FeRAS outperforms all baseline methods for all cases. As an example of the MovieLens-1M dataset with the MF model, when CR = 96.88\%, FedRAS can achieve an improvement of 7.86\% over the best HR obtained by CoLR. We can also observe that FedRAS exhibits little performance loss as the communication compression rate increases, further highlighting the effectiveness of our method in low-quality communication bandwidth scenarios.

\begin{figure}[H]
{\subfigure[ML-100K, CR = 90.63\%]{\includegraphics[width=0.495\linewidth]{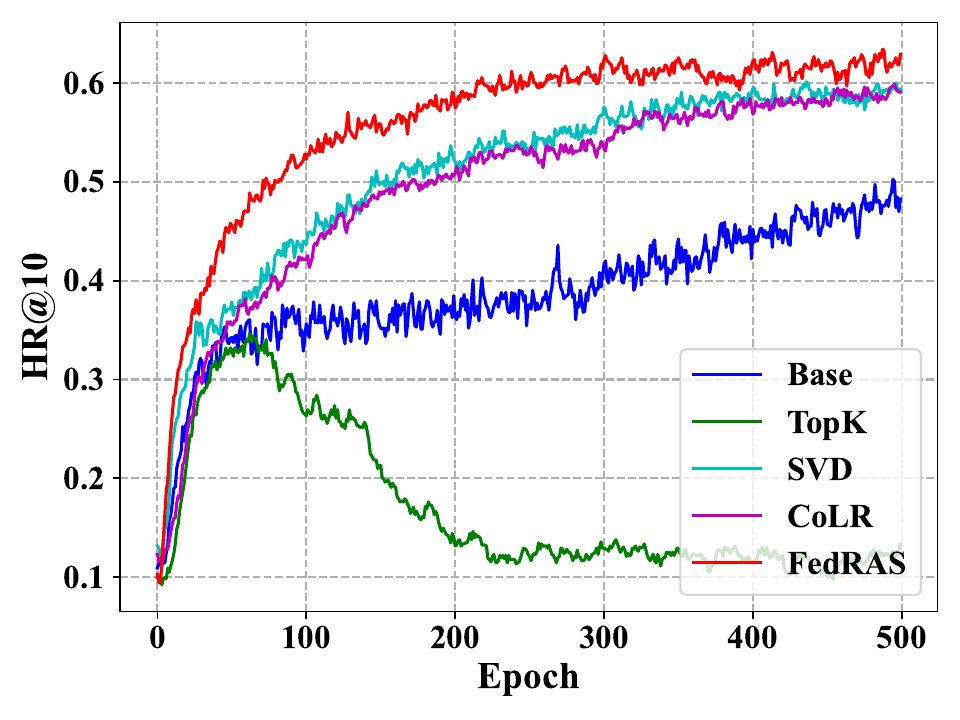}}}
{\subfigure[ML-100K, CR = 96.88\%]{\includegraphics[width=0.495\linewidth]{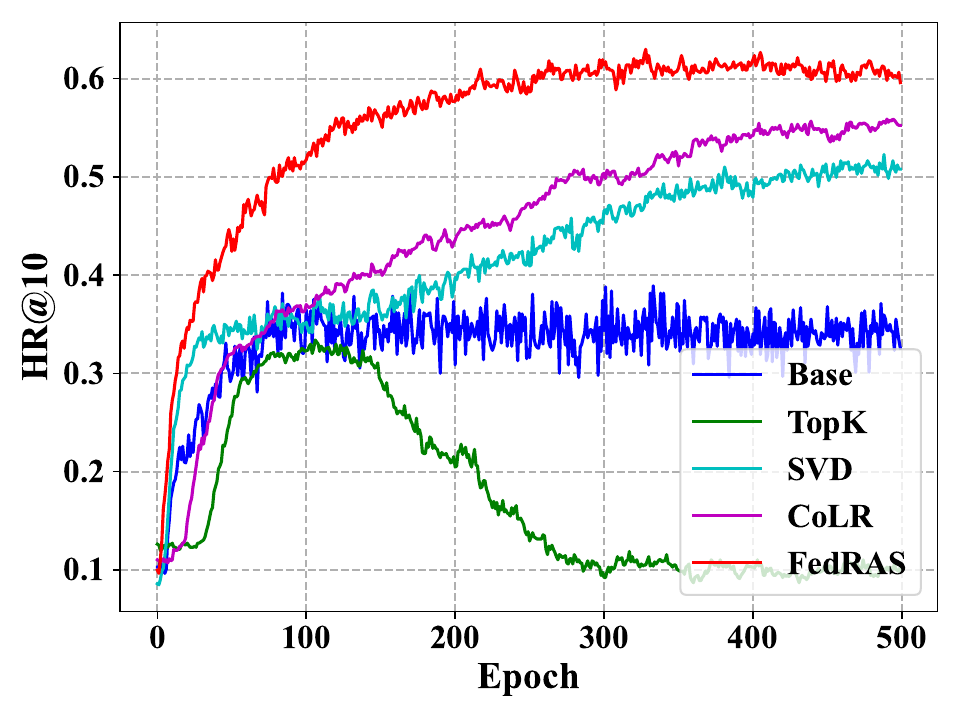}}}
{\subfigure[Lastfm-2K, CR = 90.63\%]{\includegraphics[width=0.495\linewidth]{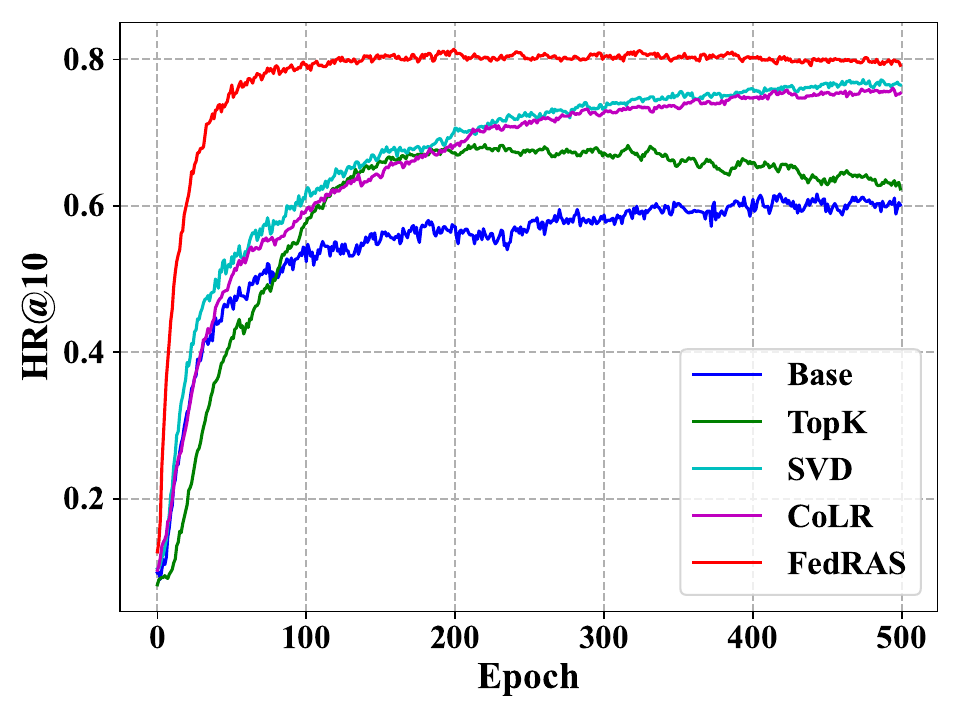}}}
{\subfigure[Lastfm-2K, CR = 96.88\%]{\includegraphics[width=0.495\linewidth]{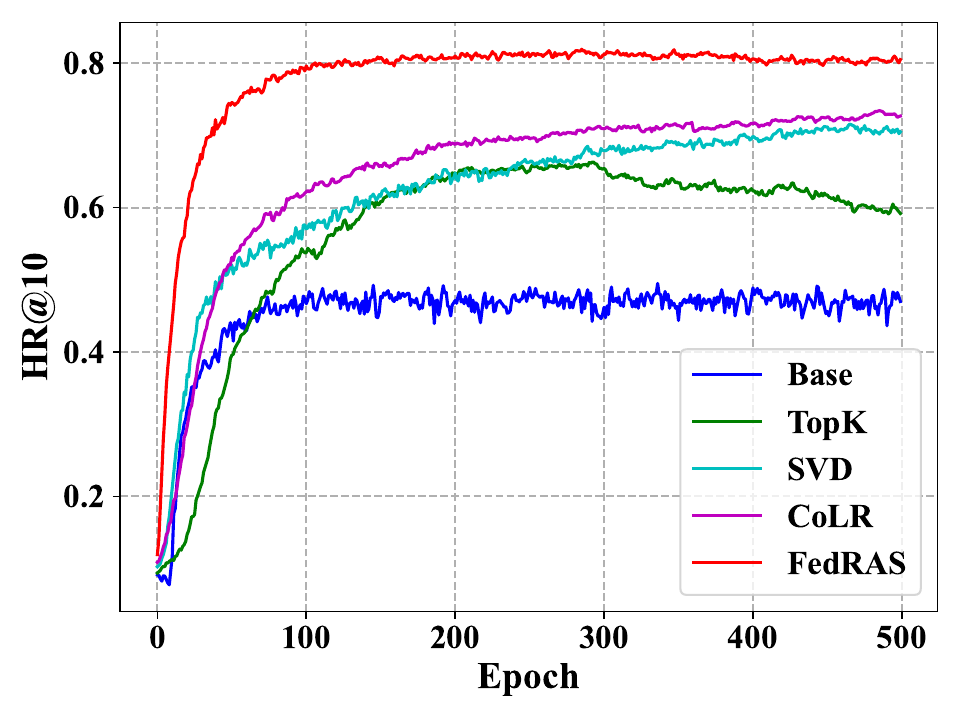}}}
\caption{Learning curves of FedRAS and other baselines.}
\label{fig: learning curves}
\end{figure}

{\bf Comparison of Model Convergence}.
Figure~\ref{fig: learning curves} presents the convergence trends of the five ML-based FL methods (including FedRAS) on the ML-100K and Lastfm-2K datasets. In each training round, we recorded the model's performance on the test dataset and used the results to plot the learning curve. 
From this figure, we can find that under the constraint of limited communication resources, FedRAS consistently outperforms other baseline models in both performance and convergence speed. This faster convergence speed further enhances the communication efficiency of FedRAS.

\subsection{Validation for Intuitions} 
To validate our intuition that the gradient sharing strategy incurs minimal information loss, we conducted an additional experiment. We adopted the MF model on the ML-1M dataset and set the compression rate to 93.75\%. Table~\ref{tab: validation} shows the information loss of baselines and of our approach, as well as compressing item embeddings by clustering (CE). We computed the information loss of compression on the client and server separately for each round and demonstrated the average. The client-side loss is defined as the mean squared error (MSE) between the aggregated compressed client gradients/item embeddings and their uncompressed counterparts, formulated as:  $\text{MSE}\left( \text{Agg}\left( \{Q_{c_u} \mid \forall u \in S \} \right), \text{Agg}\left( \{Q_{u} \mid \forall u \in S \} \right) \right)$, where $\text{Agg}(\cdot)$ denotes the server-side aggregation function, $Q_{c_u}$ and $Q_{u}$ represent the compressed and uncompressed gradients/item embeddings from client $u$ respectively. The server-side loss is defined as $\text{MSE}\left( Q, Q_c\right)$, where $Q$ and $Q_c$ denote the uncompressed and compressed global gradients/item embeddings. From Table~\ref{tab: validation}, we can find that our approach produces the minimal information loss, especially less than CE, which provides compelling evidence for the validity of our intuition. Note that there is no loss on the server for CoLR since it directly distributes the aggregated parameters to the clients. However, it still suffers from greater loss in total, which further demonstrates the superiority of our approach.

\begin{table}[H]
\caption{Information loss under different strategies.}
\label{tab: validation}
\centering
\small
\begin{tabular}{c|ccc}
\hline
Strategy & $Loss_{client}$ & $Loss_{server}$ & $Loss_{total}$ \\ \hline
CE & 2.02 & 3.66 & 5.68 \\
SVD & 7.88e-5 & 5.46 & 5.46 \\
CoLR & 3.17 & N/A & 3.17 \\
FedRAS & 1.10e-3 & 2.63 & \textbf{2.63} \\ \hline
\end{tabular}
\end{table}

\subsection{Ablation Study} 
\subsubsection{Impacts of Key Components.}
\begin{figure}[t]
{\subfigure{\includegraphics[width=0.495\linewidth]{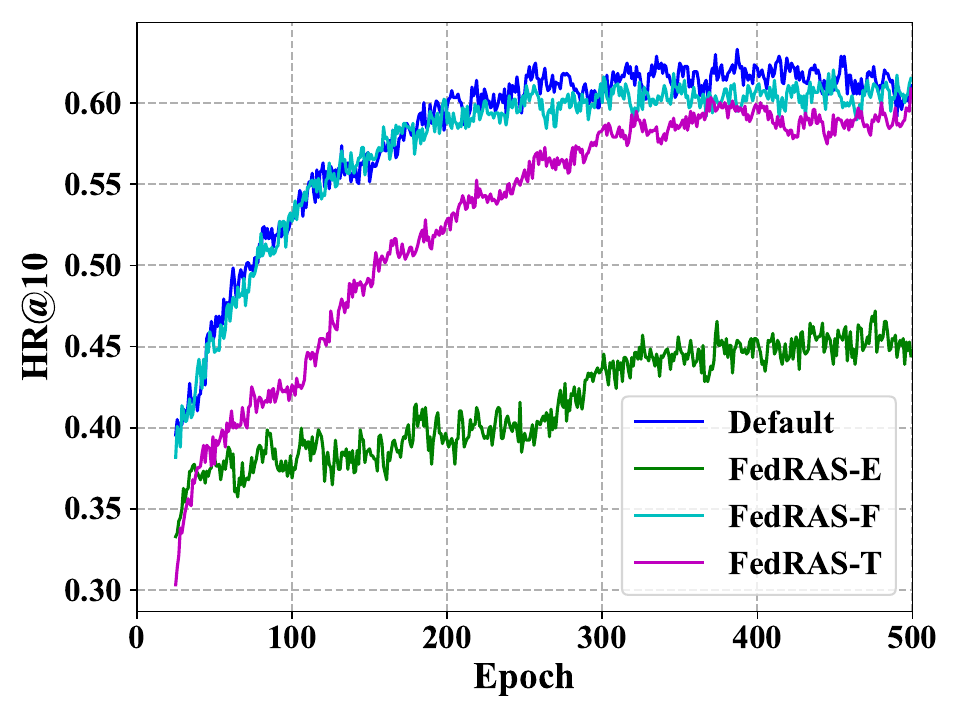}}}
{\subfigure{\includegraphics[width=0.495\linewidth]{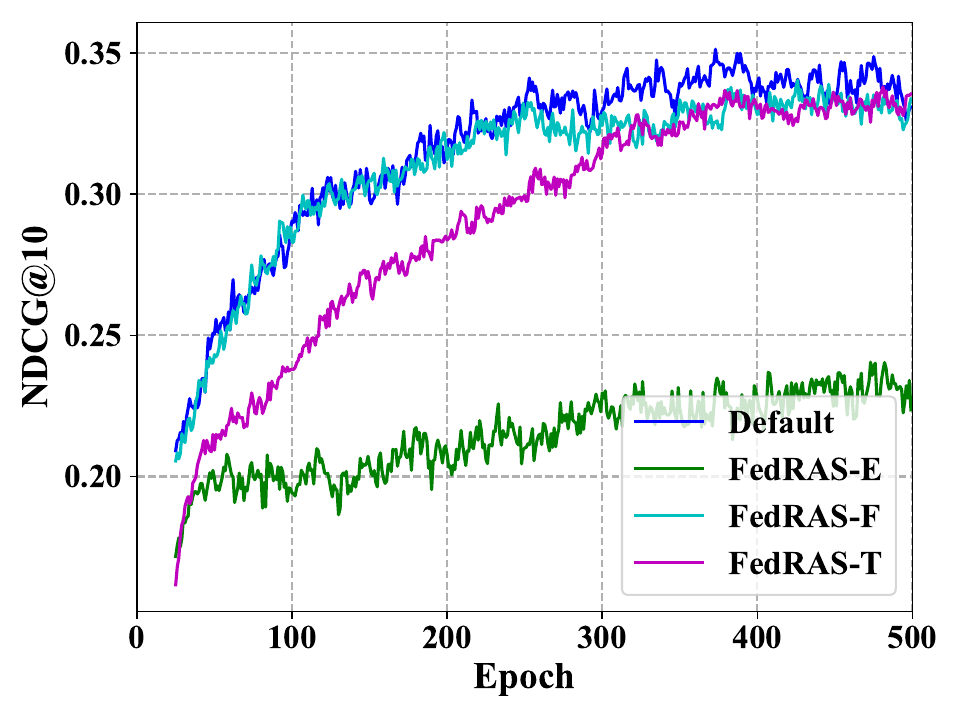}}}
\caption{Ablation study of FedRAS.}
\label{fig: ablation study}
\end{figure}

To better understand the effects produced by different key components of our method, we implemented multiple variants of FedRAS by modifying one component at a time and conducted relevant experiments on the ML-100K dataset, and then reported their results on two metrics: HR@10 and NDCG@10. The compression ratio is set to 93.75\% by default. Figure~\ref{fig: ablation study} presents the experimental results of these variants.

\textbf{Ablation of Action Sharing.}
To evaluate the effectiveness of the action sharing strategy, we developed FedRAS-E, a variant that clusters item embeddings instead of gradients. As demonstrated in Figure~\ref{fig: ablation study}, clustering raw embeddings directly leads to significantly inferior performance. This result further confirms that gradients exhibit greater robustness compared to embeddings.

\textbf{Ablation of Adaptive Gradient Clustering.}
To further validate the contribution of the adaptive clustering strategy, we introduced FedRAS-F, a variant that fixes the number of groups during downlink communication (i.e., using only K-means for gradient clustering). The result demonstrates that the adaptive clustering mechanism results in performance improvements, demonstrating the effectiveness of our method.

\textbf{Ablation of Aggregation Strategy.}
We developed FedRAS-T, which used the traditional FedAvg method as the aggregation strategy of item gradients. Compared to FedAvg, our proposed aggregation strategy achieves better performance and faster convergence, confirming that our method captures correct gradients and thus enables more accurate encoding.

\subsubsection{A Deep Analysis on Clustering.}
\begin{figure}[H]
\centering
\includegraphics[width=0.75\linewidth]{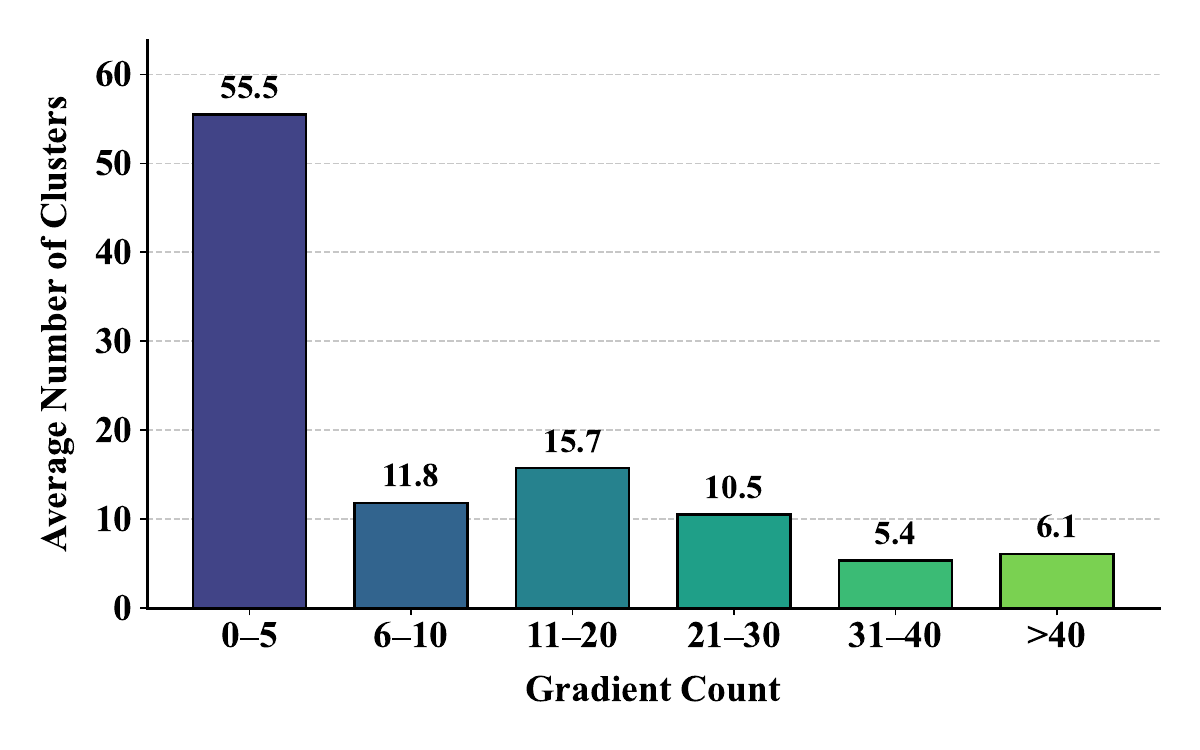}
\caption{Distribution of clusters by gradient count.}
\label{fig: clustering analysis}
\end{figure}

To analyze clustering deeply, we conducted an experiment on the ML-100K dataset, and the compression rate is set to 93.75\%. In our experiments, we recorded the average cosine similarity for each round and then computed the overall average across all rounds. The final result we obtained was \textbf{0.6631}. In addition, we counted the number of clusters with varying gradient counts and reported the average over all rounds. The results of the experiment are presented in Figure~\ref{fig: clustering analysis}. It shows that most clusters contain only a small number of gradients, resulting in relatively minor gradient disturbances after compression.

\subsection{Heterogeneous Network Bandwidth}
In this section, we explore scenarios with heterogeneous network environments and device resources. These scenarios reflect the real-world FL training process with differences in communication capabilities between clients, as exemplified in~\cite{jia2024dac, FLOverview}.
Imposing a uniform communication budget across all clients in such a heterogeneous environment becomes inefficient, as certain devices may not be able to utilize their network connections fully.

We used the ML-100K dataset for evaluation. We set up several communication bandwidth qualities and evaluated the performance under both heterogeneous and homogeneous settings. For the heterogeneous setting, we defined a range of compression rates, and the client's bandwidth limitation $e_u$ is randomly sampled from this range at the start of training and is calculated by $N*(1-CR)$. The target number of groups $C_e$ is related to the median of the compression rate range. We adjusted the fluctuation factor $\alpha$ to achieve the desired compression rate within the preset range. The results shown in Table~\ref{tab: Hete} indicate that FedRAS is effective in the context of device heterogeneity, as it can match the performance under the homogeneous setting with the same bandwidth quality.

\begin{table}[H]
\caption{Performance of FedRAS on MovieLens-100K dataset under heterogeneous and homogeneous device scenarios.}
\small
\label{tab: Hete}
\setlength{\tabcolsep}{1pt}
\begin{tabular}{c|c|c|cc}
\hline
Scenario                 & Bandwidth Quality & CR Range  & HR@10  & NDCG@10 \\ \hline
\multirow{4}{*}{Heterogeneous} & High              & 10\%-30\% & 0.6235 & 0.344   \\
                         & Medium            & 40\%-60\% & 0.6299 & 0.3464  \\
                         & Low               & 70\%-90\% & 0.6299 & 0.3361  \\
                         & Mixture           & 10\%-90\% & 0.6288 & 0.3409  \\ \hline
\multirow{4}{*}{Homogeneous}  & Excellent         & 0\%       & 0.6394 & 0.3425  \\
                         & High              & 20\%      & 0.6384 & 0.3492  \\
                         & Medium            & 50\%      & 0.6352 & 0.3472  \\
                         & Low               & 80\%      & 0.6416 & 0.347   \\ \hline
\end{tabular}
\end{table}

\begin{figure}[h]
{\subfigure[Performance of FedRAS.]
{\includegraphics[width=0.48\linewidth]{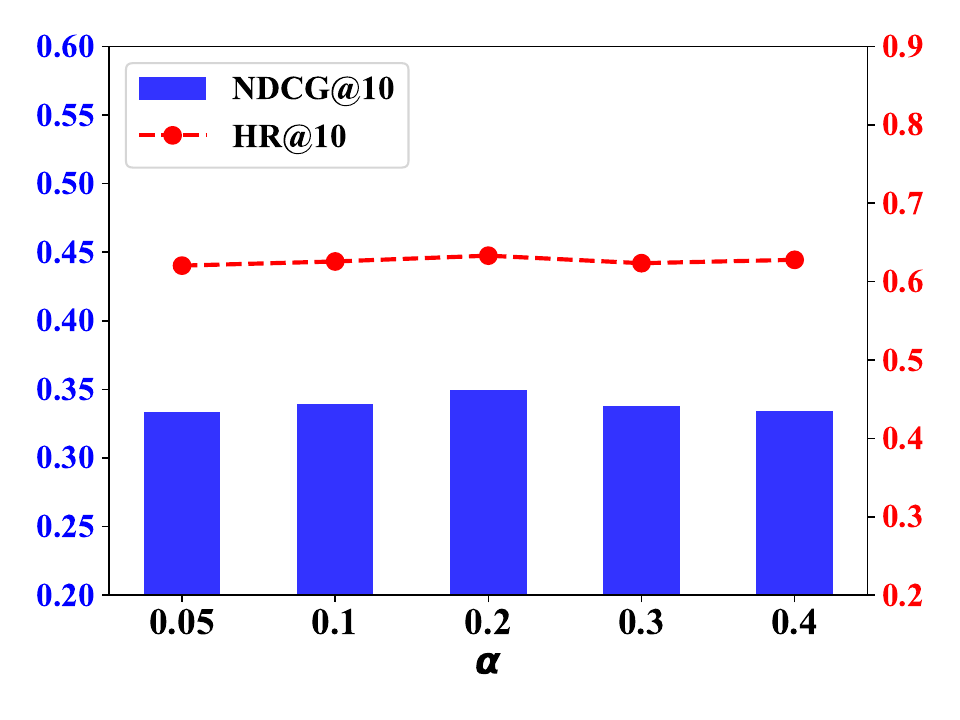}}}
{\subfigure[Average \# of groups of FedRAS.]{\includegraphics[width=0.48\linewidth]{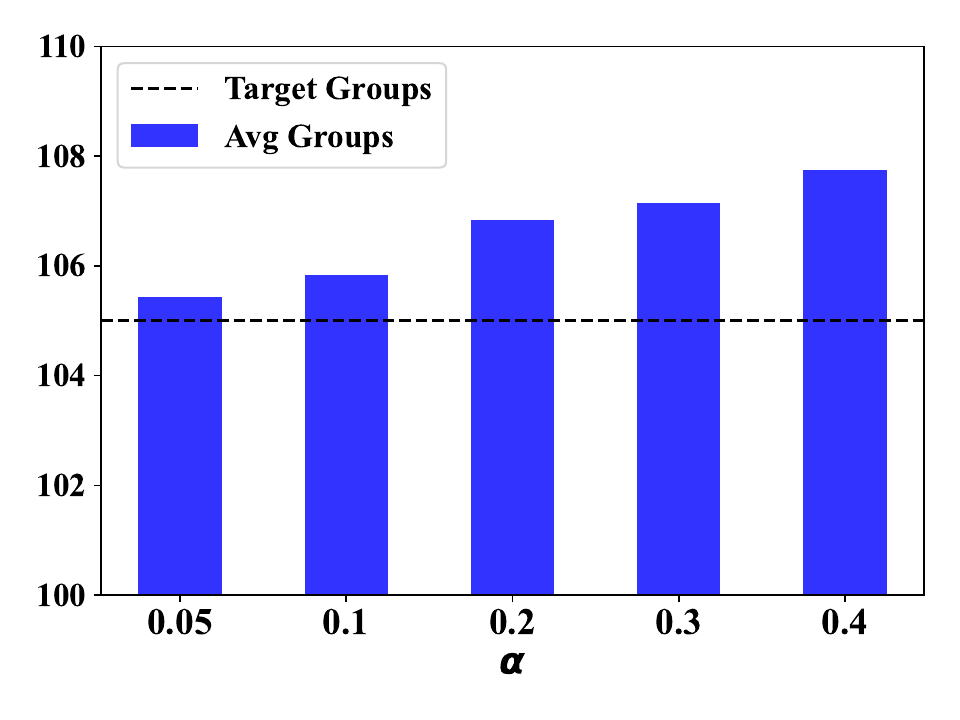}}}
\caption{Impact of fluctuation factor $\alpha$.}
\label{fig: fluctuation factor exp}
\end{figure}

\subsection{Hyperparameter Analysis}

{\bf Impact of Clustering Fluctuation Factor.}
We considered different cases of $\alpha \in \{0.05, 0.1, 0.2, 0.3, 0.4\}$, and investigated their impacts on the final performance and the average number of groups of all rounds, respectively. 
From Figure~\ref{fig: fluctuation factor exp}(a), we can find that the performance of FedRAS is relatively smooth with the change of $\alpha$, indicating that FedRAS is robust to this parameter. 
In Figure~\ref{fig: fluctuation factor exp}(b), the notion ``Target Groups'' represents the desired number of groups to achieve the preset compression rate, while the notion ``Avg Groups'' indicates the average number of groups in all rounds. 
We noticed that for each fluctuation factor $\alpha$, the average number of groups is very close to the target one. Furthermore, it can be observed that the average number of groups increases with $\alpha$, indicating that a more compact floating range limits the change per round.

\begin{figure}[H]
{\subfigure{\includegraphics[width=0.85\linewidth]{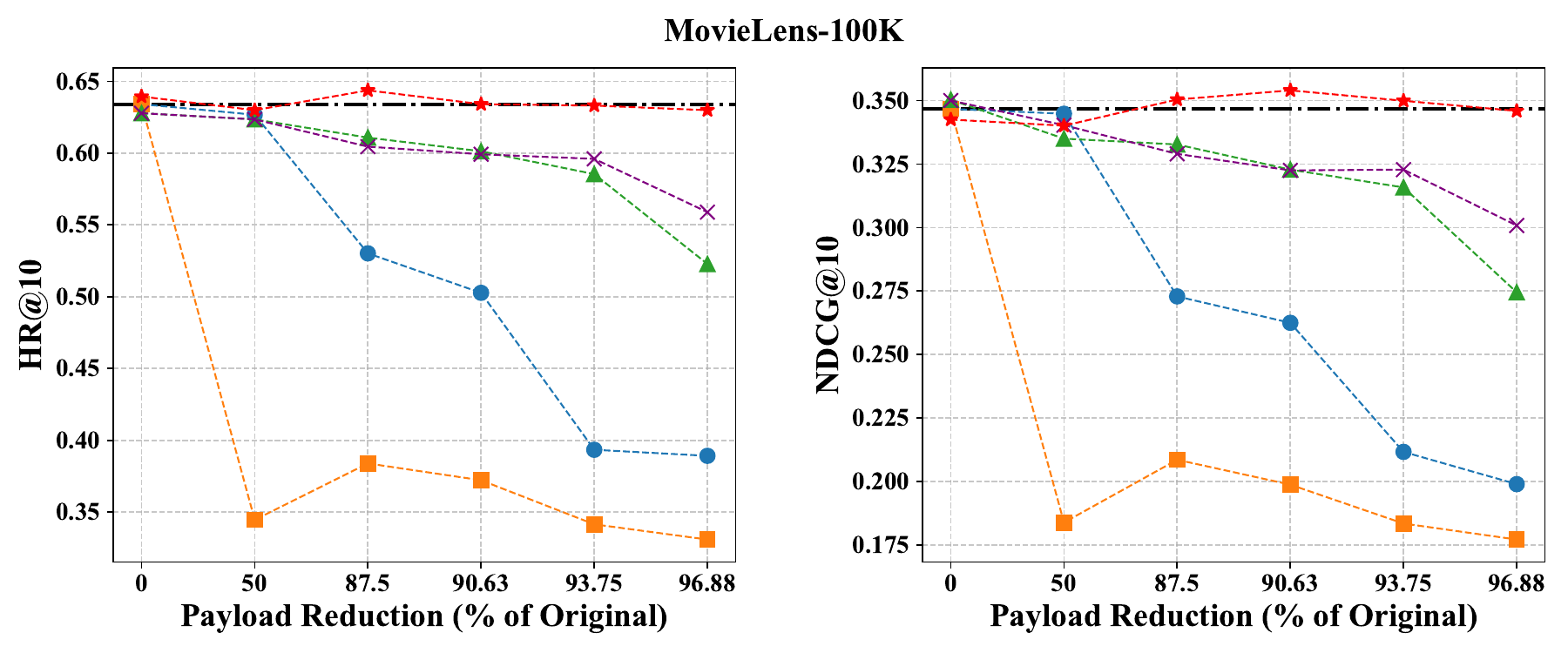}}}
{\subfigure{\includegraphics[width=0.85\linewidth]{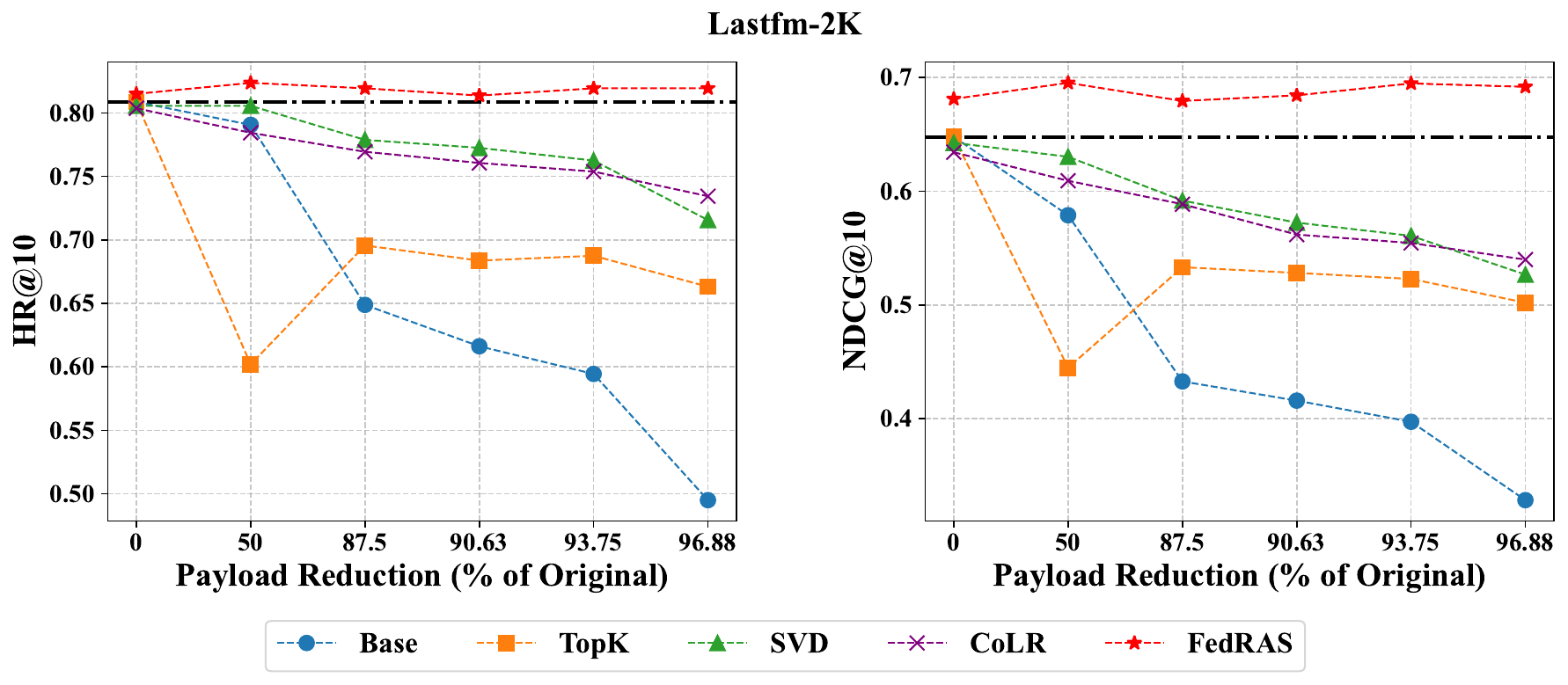}}}
\caption{Impact of communication compression rates.}
\label{fig: compression size exp}
\end{figure}

\begin{figure}[H]
{\subfigure{\includegraphics[width=0.48\linewidth]{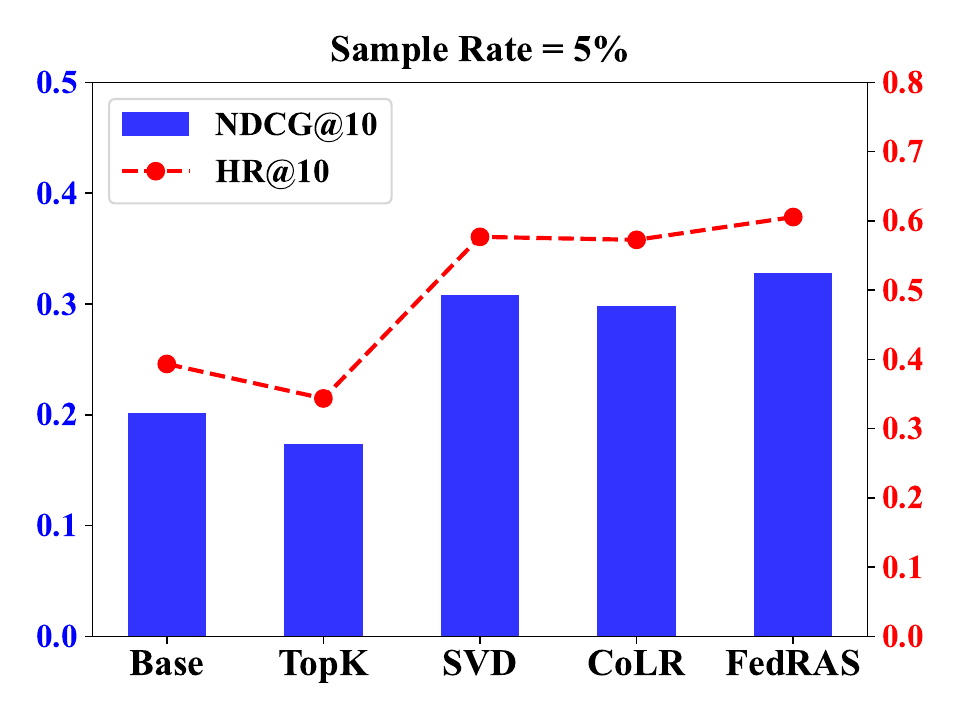}}}
{\subfigure{\includegraphics[width=0.48\linewidth]{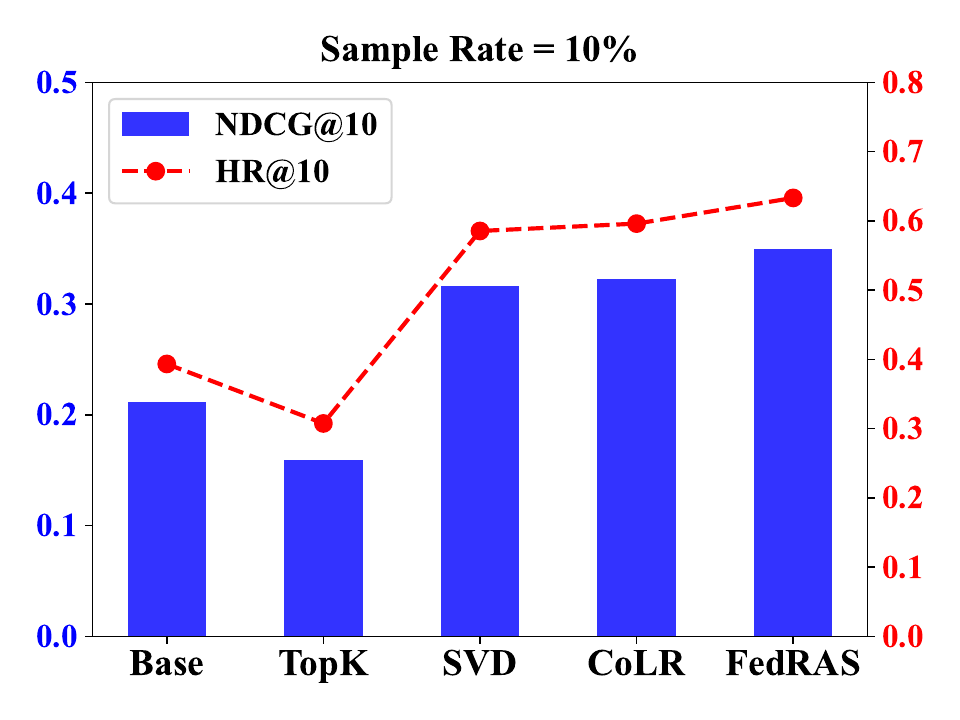}}}
{\subfigure{\includegraphics[width=0.48\linewidth]{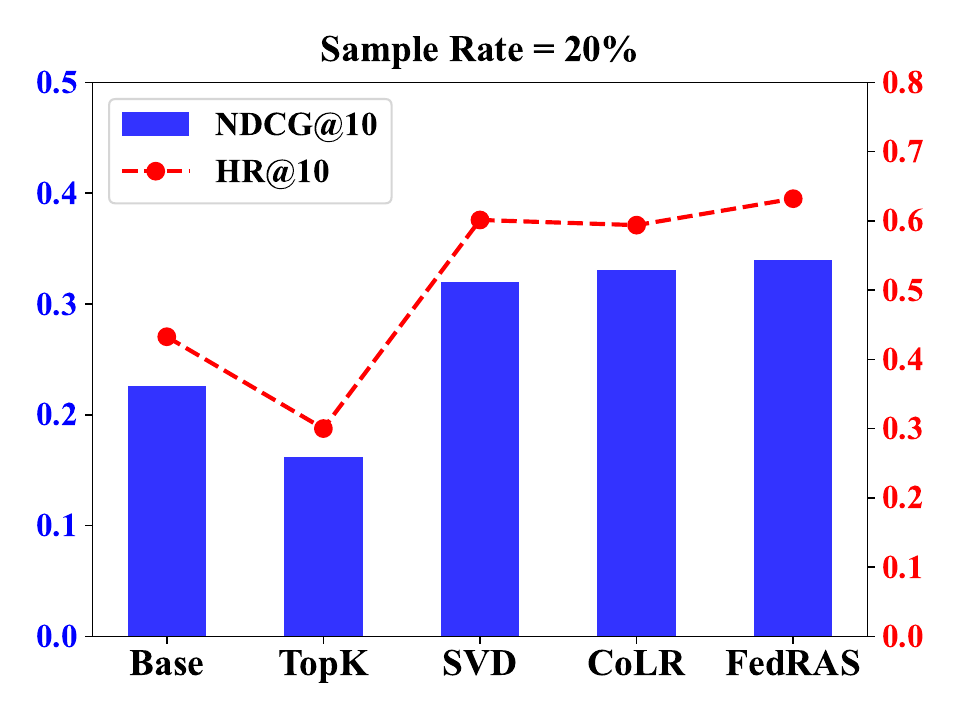}}}
{\subfigure{\includegraphics[width=0.48\linewidth]{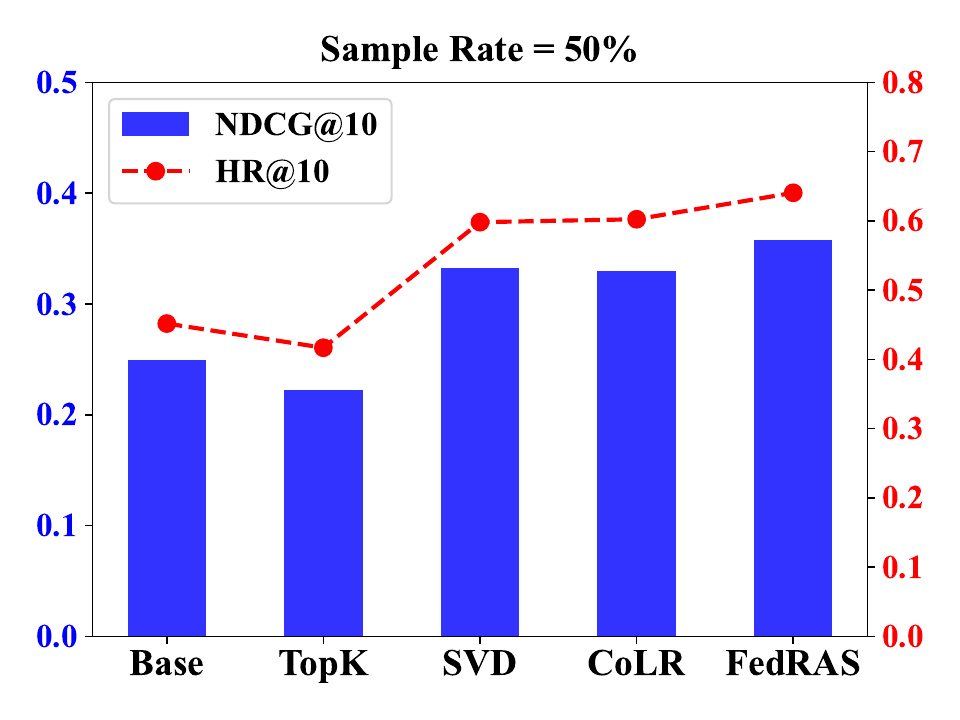}}}
\caption{Impact of sample rates of selected clients.}
\label{fig: sample ration exp}
\end{figure}

{\bf Impact of Communication Compression Rate.}
We set different compression rates to investigate the model's performance under various communication resource constraints. Figure~\ref{fig: compression size exp} presents the HR and NDCG metrics for different sizes of communication payload. The performance of the base model FedMF with the full embedding is marked as a reference for comparison.
From the figure, we can find that FedRAS achieves comparable or even superior performance across all compression rates. In contrast, most other baseline models experience a performance decline as the compression rate increases. 
This result provides additional evidence for gradient robustness: applying slight perturbations to the gradients during training can still maintain model performance. Furthermore, such mild noise may even facilitate escaping local optima, thereby enhancing overall performance.

\begin{figure}[H]
{\subfigure{\includegraphics[width=0.49\linewidth]{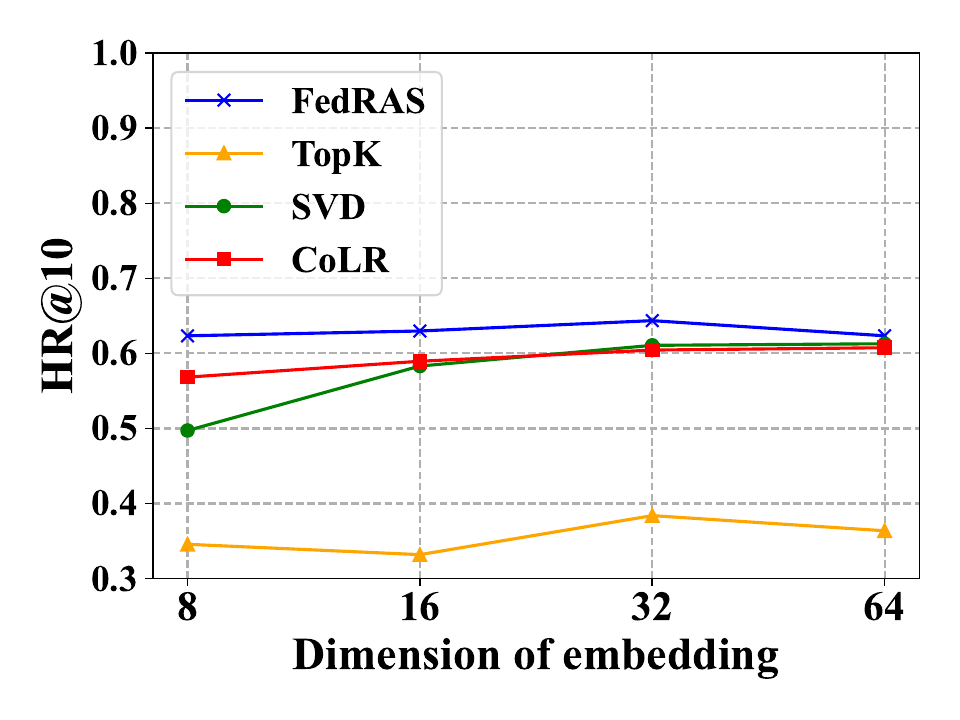}}}
{\subfigure{\includegraphics[width=0.49\linewidth]{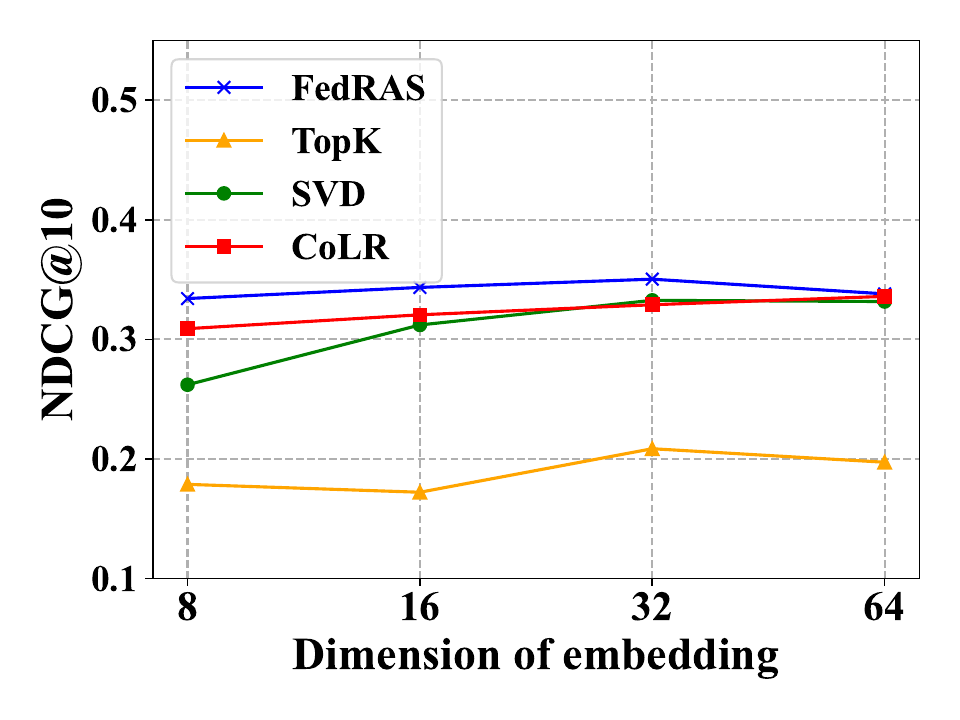}}}
\caption{Impact of embedding dimension.}
\label{fig: dimension exp}
\end{figure}

{\bf Impact of Dimension of Embedding.}
We varied the dimension of embedding in \{8, 16, 32, 64\}, and then observed the effect of this parameter on the model performance. We took the maximum compression rate of CoLR in dimension 8, i.e., 87.5\%, as the compression rate setting. The experimental results are presented in Figure \ref{fig: dimension exp}. It is clear that FedRAS shows performance comparable to or better than baselines in all dimensions.

{\bf Impact of Proportions of Selected Clients.}
To evaluate the scalability of FedRAS, we performed experiments that considered different proportions of clients participating in each training round, i.e., 5\%, 10\%, 20\% and 50\% on the ML-100K dataset with CR = 93.75\%. Figure~\ref{fig: sample ration exp} shows that FedRAS consistently outperforms the other methods in all cases.

\section{Conclusion}

In this paper, we presented a novel communication-efficient framework for federated recommendation, named FedRAS. 
Our approach reduces the high communication overhead due to massive item embeddings by an action-sharing strategy based on gradient clustering. 
In addition, to accommodate heterogeneous network environments and varying gradient distributions cross-round, we introduced a novel adaptive clustering mechanism to dynamically adjust the number of groups. 
Extensive experiments have shown superior performance gains beyond state-of-the-art baselines.



\section{Acknowledgment}

This research is supported by  Natural Science Foundation of China (62272170), Shanghai Trusted Industry Internet Software Collaborative Innovation Center, the National Research Foundation, Singapore, Cyber Security Agency of Singapore under its National Cybersecurity \&D Programme, and CyberSG R\&D Cyber Research Programme Office. Any opinions, findings and conclusions or recommendations expressed in this material are those of the author(s) and do not reflect the views of 
National Research Foundation, Singapore, Cyber Security Agency of Singapore as well as CyberSG R\&D Programme Office, Singapore. 
Zhufeng Lu, Chentao Jia, and Mingsong Chen are with  MoE Engineering Research Center of SW/HW Co-Design Technology and Application, East China Normal University.
Ming Hu (hu.ming.work@gmail.com) and Mingsong Chen (mschen@sei.ecnu.edu.cn) are the corresponding authors.

\bibliographystyle{unsrt}
\bibliography{ref}

\begin{thebibliography}{10}

\bibitem{resnick1997recommender}
Paul Resnick and Hal~R Varian.
\newblock Recommender systems.
\newblock {\em Communications of the ACM}, 40(3):56--58, 1997.

\bibitem{lu2012recommender}
Linyuan L{\"u}, Mat{\'u}{\v{s}} Medo, Chi~Ho Yeung, Yi-Cheng Zhang, Zi-Ke Zhang, and Tao Zhou.
\newblock Recommender systems.
\newblock {\em Physics Reports}, 519(1):1--49, 2012.

\bibitem{barkan2024counterfactual}
Oren Barkan, Veronika Bogina, Liya Gurevitch, Yuval Asher, and Noam Koenigstein.
\newblock A counterfactual framework for learning and evaluating explanations for recommender systems.
\newblock In {\em Proceedings of the ACM on Web Conference (WWW)}, pages 3723--3733, 2024.

\bibitem{jung2023dual}
Heesoo Jung, Sangpil Kim, and Hogun Park.
\newblock Dual policy learning for aggregation optimization in graph neural network-based recommender systems.
\newblock In {\em Proceedings of the ACM Web Conference (WWW)}, pages 1478--1488, 2023.

\bibitem{GARCIASANCHEZ2020102153}
Francisco García-Sánchez, Ricardo Colomo-Palacios, and Rafael Valencia-García.
\newblock A social-semantic recommender system for advertisements.
\newblock {\em Information Processing \& Management}, 57(2):102153, 2020.

\bibitem{4280214}
Kangning Wei, Jinghua Huang, and Shaohong Fu.
\newblock A survey of e-commerce recommender systems.
\newblock In {\em Proceedings of International Conference on Service Systems and Service Management}, pages 1--5, 2007.

\bibitem{bonta2022california}
Rob Bonta.
\newblock California consumer privacy act.
\newblock {\em Retrieved from State of California Department of Justice}, 2022.

\bibitem{voigt2017eu}
Paul Voigt and Axel Von~dem Bussche.
\newblock {\em The {EU} general data protection regulation ({GDPR}) A Practical Guide}.
\newblock Springer Cham, 2017.

\bibitem{fedavg}
Brendan McMahan, Eider Moore, Daniel Ramage, Seth Hampson, and Blaise~Aguera y~Arcas.
\newblock Communication-efficient learning of deep networks from decentralized data.
\newblock In {\em Proceedings of the Artificial Intelligence and Statistics (AISTATS)}, pages 1273--1282, 2017.

\bibitem{huang2024federated}
Wenke Huang, Mang Ye, Zekun Shi, Guancheng Wan, He~Li, Bo~Du, and Qiang Yang.
\newblock Federated learning for generalization, robustness, fairness: A survey and benchmark.
\newblock {\em IEEE Transactions on Pattern Analysis and Machine Intelligence}, 46(12):9387--9406, 2024.

\bibitem{hu2024fedmut}
Ming Hu, Yue Cao, Anran Li, Zhiming Li, Chengwei Liu, Tianlin Li, Mingsong Chen, and Yang Liu.
\newblock Fedmut: Generalized federated learning via stochastic mutation.
\newblock In {\em Proceedings of the AAAI Conference on Artificial Intelligence}, volume~38, pages 12528--12537, 2024.

\bibitem{qi2023cross}
Zhuang Qi, Lei Meng, Zitan Chen, Han Hu, Hui Lin, and Xiangxu Meng.
\newblock Cross-silo prototypical calibration for federated learning with non-iid data.
\newblock In {\em Proceedings of the 31st ACM International Conference on Multimedia}, pages 3099--3107, 2023.

\bibitem{hu2024aggregation}
Ming Hu, Zhihao Yue, Xiaofei Xie, Cheng Chen, Yihao Huang, Xian Wei, Xiang Lian, Yang Liu, and Mingsong Chen.
\newblock Is aggregation the only choice? federated learning via layer-wise model recombination.
\newblock In {\em Proceedings of the ACM SIGKDD Conference on Knowledge Discovery and Data Mining (KDD)}, pages 1096--1107, 2024.

\bibitem{xia2025multisfl}
Zeke Xia, Ming Hu, Dengke Yan, Ruixuan Liu, Anran Li, Xiaofei Xie, and Mingsong Chen.
\newblock Multisfl: Towards accurate split federated learning via multi-model aggregation and knowledge replay.
\newblock In {\em Proceedings of the AAAI Conference on Artificial Intelligence}, volume~39, pages 914--922, 2025.

\bibitem{liao2024swiss}
Tianchi Liao, Lele Fu, Jialong Chen, Zhen Wang, Zibin Zheng, and Chuan Chen.
\newblock A swiss army knife for heterogeneous federated learning: Flexible coupling via trace norm.
\newblock {\em Advances in Neural Information Processing Systems}, 37:139886--139911, 2024.

\bibitem{li2024joint}
Anran Li, Guangjing Wang, Ming Hu, Jianfei Sun, Lan Zhang, Luu~Anh Tuan, and Han Yu.
\newblock Joint client-and-sample selection for federated learning via bi-level optimization.
\newblock {\em IEEE Transactions on Mobile Computing}, 2024.

\bibitem{yang2024improved}
Banglie Yang, Linyu Zhu, Cheng Dai, Sahil Garg, and Georges Kaddoum.
\newblock An improved reconstruction based multi-attribute contrastive learning for digital twin-enabled industrial system.
\newblock {\em IEEE Internet of Things Journal}, 2024.

\bibitem{muhammad2020fedfast}
Khalil Muhammad, Qinqin Wang, Diarmuid O'Reilly-Morgan, Elias Tragos, Barry Smyth, Neil Hurley, James Geraci, and Aonghus Lawlor.
\newblock Fedfast: Going beyond average for faster training of federated recommender systems.
\newblock In {\em Proceedings of the ACM SIGKDD Conference on Knowledge Discovery and Data Mining (KDD)}, pages 1234--1242, 2020.

\bibitem{wang2022fast}
Qinyong Wang, Hongzhi Yin, Tong Chen, Junliang Yu, Alexander Zhou, and Xiangliang Zhang.
\newblock Fast-adapting and privacy-preserving federated recommender system.
\newblock {\em The VLDB Journal}, 31(5):877--896, 2022.

\bibitem{yang2020federated}
Liu Yang, Ben Tan, Vincent~W Zheng, Kai Chen, and Qiang Yang.
\newblock Federated recommendation systems.
\newblock In {\em Federated Learning: Privacy and Incentive}, pages 225--239. Springer, 2020.

\bibitem{wang2021poi}
Li-e Wang, Yihui Wang, Yan Bai, Peng Liu, and Xianxian Li.
\newblock Poi recommendation with federated learning and privacy preserving in cross domain recommendation.
\newblock In {\em IEEE INFOCOM 2021-IEEE Conference on Computer Communications Workshops (INFOCOM WKSHPS)}, pages 1--6. IEEE, 2021.

\bibitem{hu2023gitfl}
Ming Hu, Zeke Xia, Dengke Yan, Zhihao Yue, Jun Xia, Yihao Huang, Yang Liu, and Mingsong Chen.
\newblock Gitfl: Uncertainty-aware real-time asynchronous federated learning using version control.
\newblock In {\em 2023 IEEE Real-Time Systems Symposium (RTSS)}, pages 145--157. IEEE, 2023.

\bibitem{cui2022optimizing}
Yangguang Cui, Kun Cao, Junlong Zhou, and Tongquan Wei.
\newblock Optimizing training efficiency and cost of hierarchical federated learning in heterogeneous mobile-edge cloud computing.
\newblock {\em IEEE transactions on computer-aided design of integrated circuits and systems}, 42(5):1518--1531, 2022.

\bibitem{wang2023dafkd}
Haozhao Wang, Yichen Li, Wenchao Xu, Ruixuan Li, Yufeng Zhan, and Zhigang Zeng.
\newblock Dafkd: Domain-aware federated knowledge distillation.
\newblock In {\em Proceedings of the IEEE/CVF conference on Computer Vision and Pattern Recognition}, pages 20412--20421, 2023.

\bibitem{flexfl}
Zekai Chen, Chentao Jia, Ming Hu, Xiaofei Xie, Anran Li, and Mingsong Chen.
\newblock Flexfl: Heterogeneous federated learning via apoz-guided flexible pruning in uncertain scenarios.
\newblock {\em IEEE Transactions on Computer-Aided Design of Integrated Circuits and Systems (TCAD)}, 43(11):4069--4080, 2024.

\bibitem{gao2024nebulafl}
Fei Gao, Ming Hu, Zhiyu Xie, Peichang Shi, Xiaofei Xie, Guodong Yi, and Huaimin Wang.
\newblock Nebulafl: Effective asynchronous federated learning for jointcloud computing.
\newblock {\em arXiv preprint arXiv:2412.04868}, 2024.

\bibitem{FedMF}
Di~Chai, Leye Wang, Kai Chen, and Qiang Yang.
\newblock Secure federated matrix factorization.
\newblock {\em IEEE Intelligent Systems}, 36(5):11--20, 2020.

\bibitem{JointRec}
Sijing Duan, Deyu Zhang, Yanbo Wang, Lingxiang Li, and Yaoxue Zhang.
\newblock {JointRec}: A deep-learning-based joint cloud video recommendation framework for mobile iot.
\newblock {\em IEEE Internet of Things Journal}, 7(3):1655--1666, 2019.

\bibitem{CoLR}
Ngoc{-}Hieu Nguyen, Tuan{-}Anh Nguyen, Tuan Nguyen, Vu~Tien Hoang, Dung~D. Le, and Kok{-}Seng Wong.
\newblock Towards efficient communication and secure federated recommendation system via low-rank training.
\newblock In {\em Proceedings of the ACM on Web Conference (WWW)}, pages 3940--3951, 2024.

\bibitem{FCF-BST}
Farwa~K Khan, Adrian Flanagan, Kuan~Eeik Tan, Zareen Alamgir, and Muhammad Ammad-Ud-Din.
\newblock A payload optimization method for federated recommender systems.
\newblock In {\em Proceedings of the ACM Conference on Recommender Systems}, pages 432--442, 2021.

\bibitem{kmeans}
Tapas Kanungo, David~M Mount, Nathan~S Netanyahu, Christine~D Piatko, Ruth Silverman, and Angela~Y Wu.
\newblock An efficient k-means clustering algorithm: Analysis and implementation.
\newblock {\em IEEE transactions on pattern analysis and machine intelligence}, 24(7):881--892, 2002.

\bibitem{hu2024fedcross}
Ming Hu, Peiheng Zhou, Zhihao Yue, Zhiwei Ling, Yihao Huang, Anran Li, Yang Liu, Xiang Lian, and Mingsong Chen.
\newblock Fedcross: Towards accurate federated learning via multi-model cross-aggregation.
\newblock In {\em 2024 IEEE 40th International Conference on Data Engineering (ICDE)}, pages 2137--2150. IEEE, 2024.

\bibitem{xia2024cabafl}
Zeke Xia, Ming Hu, Dengke Yan, Xiaofei Xie, Tianlin Li, Anran Li, Junlong Zhou, and Mingsong Chen.
\newblock Cabafl: Asynchronous federated learning via hierarchical cache and feature balance.
\newblock {\em IEEE Transactions on Computer-Aided Design of Integrated Circuits and Systems}, 43(11):4057--4068, 2024.

\bibitem{FCF}
Muhammad Ammad-Ud-Din, Elena Ivannikova, Suleiman~A Khan, Were Oyomno, Qiang Fu, Kuan~Eeik Tan, and Adrian Flanagan.
\newblock Federated collaborative filtering for privacy-preserving personalized recommendation system.
\newblock {\em arXiv preprint arXiv:1901.09888}, 2019.

\bibitem{FedRec}
Guanyu Lin, Feng Liang, Weike Pan, and Zhong Ming.
\newblock Fedrec: Federated recommendation with explicit feedback.
\newblock {\em IEEE Intelligent Systems}, 36(5):21--30, 2020.

\bibitem{MetaMF}
Yujie Lin, Pengjie Ren, Zhumin Chen, Zhaochun Ren, Dongxiao Yu, Jun Ma, Maarten~de Rijke, and Xiuzhen Cheng.
\newblock Meta matrix factorization for federated rating predictions.
\newblock In {\em Proceedings of the International ACM SIGIR Conference on Research and Development in Information Retrieval}, pages 981--990, 2020.

\bibitem{FedNCF}
Vasileios Perifanis and Pavlos~S Efraimidis.
\newblock Federated neural collaborative filtering.
\newblock {\em Knowledge-Based Systems}, 242:108441, 2022.

\bibitem{NCF}
Xiangnan He, Lizi Liao, Hanwang Zhang, Liqiang Nie, Xia Hu, and Tat-Seng Chua.
\newblock Neural collaborative filtering.
\newblock {\em arXiv preprint arXiv:1708.05031}, 2017.

\bibitem{FedPerGNN}
Chuhan Wu, Fangzhao Wu, Lingjuan Lyu, Tao Qi, Yongfeng Huang, and Xing Xie.
\newblock A federated graph neural network framework for privacy-preserving personalization.
\newblock {\em Nature Communications}, 13(1):3091, 2022.

\bibitem{ML}
F.~Maxwell Harper and Joseph~A. Konstan.
\newblock The movielens datasets: History and context.
\newblock {\em ACM Transactions on Intelligent Systems and Technology}, 5(4):19:1--19:19, 2016.

\bibitem{Lastfm}
Iv{\'a}n Cantador, Peter Brusilovsky, and Tsvi Kuflik.
\newblock Second workshop on information heterogeneity and fusion in recommender systems (hetrec2011).
\newblock In {\em Proceedings of the fifth ACM conference on Recommender systems}, pages 387--388, 2011.

\bibitem{jia2024dac}
Chentao Jia, Ming Hu, Zekai Chen, Yanxin Yang, Xiaofei Xie, Yang Liu, and Mingsong Chen.
\newblock Adaptivefl: Adaptive heterogeneous federated learning for resource-constrained aiot systems.
\newblock In {\em Proceedings of The Chips To Systems Conference (DAC)}, pages 1--6, 2024.

\bibitem{FLOverview}
Tian Li, Anit~Kumar Sahu, Ameet Talwalkar, and Virginia Smith.
\newblock Federated learning: Challenges, methods, and future directions.
\newblock {\em IEEE Signal Processing Magazine}, 37(3):50--60, 2020.

\bibitem{LightFR}
Honglei Zhang, Fangyuan Luo, Jun Wu, Xiangnan He, and Yidong Li.
\newblock Lightfr: Lightweight federated recommendation with privacy-preserving matrix factorization.
\newblock {\em ACM Transactions on Information Systems}, 41(4):1--28, 2023.

\bibitem{li2020sampling}
Dong Li, Ruoming Jin, Jing Gao, and Zhi Liu.
\newblock On sampling top-k recommendation evaluation.
\newblock In {\em Proceedings of the 26th ACM SIGKDD International Conference on Knowledge Discovery \& Data Mining}, pages 2114--2124, 2020.

\end{thebibliography}

\clearpage
\appendix
\balance
\section{APPENDIX}

\subsection{Comparison with More Baselines}
We conducted experiments with more baselines to validate the effectiveness of FedRAS. The baselines we adopted are as follows:
\begin{itemize}

\item[$\bullet$] \textbf{RCR} is a vanilla method to reduce communication overhead by reducing communication rounds. We reduced the number of training rounds according to the compression rates, with the full training consisting of 500 rounds.

\item[$\bullet$] \textbf{RSC} is another vanilla method to reduce communication overhead by reducing the number of selected clients per round according to the compression rates, with the full training consisting of 10\% of users per round.

\item[$\bullet$] \textbf{LightFR~\cite{LightFR}} is a federated recommendation framework that uses high-quality binary codes by exploiting the learning-to-hash technique and devises a discrete optimization algorithm to collaboratively train the parameters of the model between the server and the clients.

\item[$\bullet$] \textbf{JointRec~\cite{JointRec}} combines low rank matrix factorization and an 8-bit quantization method to reduce communication costs and network bandwidth.

\end{itemize}

\begin{table}[H]
\caption{Performance comparison with new baselines.}
\label{tab: moreExp}
\centering
\small
\begin{tabular}{c|ccc}
\hline
Method & CR=90.63\% & CR=93.75\% & CR=96.88\% \\ \hline
RCR & 0.3733/0.2050 & 0.3340/0.1789 & 0.2768/0.1497 \\
RSC & 0.5217/0.2724 & 0.4836/0.2444 & 0.3796/0.1920 \\
LightFR & 0.4788/0.2854 & 0.4645/0.2742 & 0.4639/0.2653 \\
JointRec & 0.6246/0.3447 & 0.6076/0.3283 & 0.5907/0.3210 \\
FedRAS & \textbf{0.6341/0.3540} & \textbf{0.6331/0.3499} & \textbf{0.6299/0.3459} \\ \hline
\end{tabular}
\end{table}

We conducted the experiment on the base model MF and the dataset ML-100K. The experimental settings are the same as in Section~\ref{sec: expr}. From Table~\ref{tab: moreExp}, we can find that our approach performs best. Meanwhile, as the compression rate increases, the performance gap between FedRAS and the baseline continues to widen. For instance, at CR=90.63\%, FedRAS achieves a 0.0095 improvement in HR over JointRec, while at CR=96.88\%, the improvement in HR reaches 0.0392. This further demonstrates the adaptability of our method to low-bandwidth scenarios.

\subsection{Analysis on Computational Overhead}
To demonstrate the computational overhead of clustering, we contrast the computation and communication time of uncompressed and compressed scenarios under different baselines. On a Lastfm‑2K-sized task (12454 items, 64‑dimensional 32‑bit embeddings), the size of the gradient matrix is about 3.04MB. At the 0.75MB/second downlink transmission speed reported in~\cite{CoLR}, it would take about 4.05 seconds for the server to transmit the uncompressed gradients.

We then calculated the computation time under different baselines, and compared the total time consumption. We treated our experimental environment as a server, and recorded the time it took to perform a single compression. The compression rate is set to 93.75\%. For k-means, we used the function provided by the Scikit-Learn module. For SVD, we used the function provided by the Numpy module. 

\begin{table}[H]
\caption{Analysis on computational overhead.}
\label{tab: comp}
\centering
\footnotesize
\begin{tabular}{c|c|c|c|c}
\hline
Method & Comm. Time & Comp. Time & Total Time & Performance \\ \hline
w/o compression & 4.05s & 0s & 4.05s & 0.6341/0.3467 \\
TopK & 0.25s & 7e-4s & 0.25s & 0.3415/0.1835 \\
SVD & 0.25s & 5.70s & 5.95s & 0.5854/0.3158 \\
FedRAS & 0.25s & 0.82s & 1.07s & 0.6331/0.3499 \\ \hline
\end{tabular}
\end{table}

Table~\ref{tab: comp} presents the experimental results. In the table, ``Comm'' denotes communication and ``Comp'' denotes computation. The performance column contains two metrics, HR@10 and NDCG@10, separated by a forward slash (/). The results show that, compared to the total time cost of no compression, compression brings great benefits. Our algorithm performed noticeably faster than SVD while exceeding its performance. While Top‑K required little computation, it suffers from significant performance degradation.

\begin{table}[H]
\caption{Experimental results on the new evaluation method.}
\label{tab: full eval}
\centering
\small
\begin{tabular}{c|ccc}
\hline
Method & CR=90.63\% & CR=93.75\% & CR=96.88\% \\ \hline
Base & 0.1144/0.0581 & 0.1050/0.0566 & 0.0663/0.0241 \\
TopK & 0.1781/0.1166 & 0.1681/0.1119 & 0.1562/0.1079 \\
SVD & 0.1812/0.1018 & 0.1669/0.0780 & 0.1444/0.0621 \\
CoLR & 0.1537/0.0818 & 0.1475/0.0792 & 0.1450/0.0758 \\
FedRAS & \textbf{0.3312/0.1409} & \textbf{0.3250/0.1328} & \textbf{0.2963/0.1274} \\ \hline
\end{tabular}
\end{table}

\subsection{Experiments on a New Evaluation Method}
Previous study~\cite{li2020sampling} has indicated that sampling-based evaluation methods may introduce bias. 
To assess the performance of our method more comprehensively, we employ a full-set evaluation approach, i.e., treating all non-interacted items as the candidate set rather than negative sampling. 

We conducted the experiment on the base model MF and the dataset Lastfm-2K. The experimental settings are the same as in Section~\ref{sec: expr}. 
Table~\ref{tab: full eval} shows the experimental results.

\end{document}